\documentclass[10pt,journal,cspaper,compsoc]{IEEEtran}
\ifCLASSINFOpdf
\else
\fi
\usepackage{amssymb}

\usepackage{epsfig}
\usepackage{epstopdf}
\usepackage{subfigure}
\usepackage{amsmath}
\usepackage{booktabs}

\usepackage{times}
\usepackage{graphicx}
\usepackage{amssymb}

\usepackage{color}
\usepackage{cite}
\usepackage{psfig}
\usepackage{algorithm}
\usepackage{algorithmic}
\usepackage{graphicx}
\usepackage{cases}
\usepackage{multirow}

\usepackage{psfig}
\usepackage{algorithm}
\usepackage{algorithmic}
\usepackage{graphicx}
\usepackage{amsmath}
\usepackage{booktabs}
\usepackage{subfigure}
\usepackage{cases}
\usepackage{enumerate}
\usepackage{wrapfig}
\usepackage{picinpar}
\usepackage{pbox}

\usepackage{hyperref}
\usepackage{subfigure}
\usepackage{amsmath}
\usepackage{pifont}
\newcommand{\xmark}{\ding{55}}%

\newcommand{\thickhline}{%
	\noalign {\ifnum 0=`}\fi \hrule height 1pt
	\futurelet \reserved@a \@xhline
}

\begin{document}
%
\title{Geometric and Learning-based Mesh Denoising: A Comprehensive Survey}

\author{
\IEEEauthorblockN{Honghua Chen, Mingqiang Wei and Jun Wang} \\
\thanks{H. Chen, M. Wei, and J. Wang are all from Nanjing University of Aeronautics and Astronautics. \newline
Email: chenhonghuacn@gmail.com, mingqiang.wei@gmail.com, davis.wjun@gmail.com.  \newline
 Corresponding author: J. Wang}
}

%



\IEEEtitleabstractindextext{%
\begin{abstract}
 	Mesh denoising is a fundamental problem in digital geometry processing. It seeks to remove surface noise, while preserving surface intrinsic signals as accurately as possible. While the traditional wisdom has been built upon specialized priors to smooth surfaces, learning-based approaches are making their debut with great success in generalization and automation. In this work, we provide a comprehensive review of the advances in mesh denoising, containing both traditional geometric approaches and recent learning-based methods. First, to familiarize readers with the denoising tasks, we summarize four common issues in mesh denoising. We then provide two categorizations of the existing denoising methods. Furthermore, three important categories, including optimization-, filter-, and data-driven-based techniques, are introduced and analyzed in detail, respectively. Both qualitative and quantitative comparisons are illustrated, to demonstrate the effectiveness of the state-of-the-art denoising methods. Finally, potential directions of future work are pointed out to solve the common problems of these approaches. A mesh denoising benchmark is also built in this work, and future researchers will easily and conveniently evaluate their methods with the state-of-the-art approaches.
\end{abstract}

}

\maketitle

\IEEEdisplaynontitleabstractindextext

%
\IEEEpeerreviewmaketitle

%
%
%
%

\section{Introduction}
Recent advances in scanning devices and improvements in techniques that generate and synthesize 3D shapes have made 3D mesh surfaces widespread in various fields, including computer graphics, virtual reality, reverse engineering, etc. However, during the acquisition of a 3D mesh surface, various noises from uncertain sources inevitably creep in and strongly tamper the data quality. This hinders many downstream applications, like model reconstruction, visualization, numerical simulation, etc. Consequently, it is desirable to be able to acquire a high-fidelity 3D model by mesh denoising, when considering the above applications. 

\begin{figure}
	\centering
	\subfigure[Denoising \cite{botsch2010polygon}]{
		\label{fig:definition:denoising} 
		\includegraphics[width=1.7in]{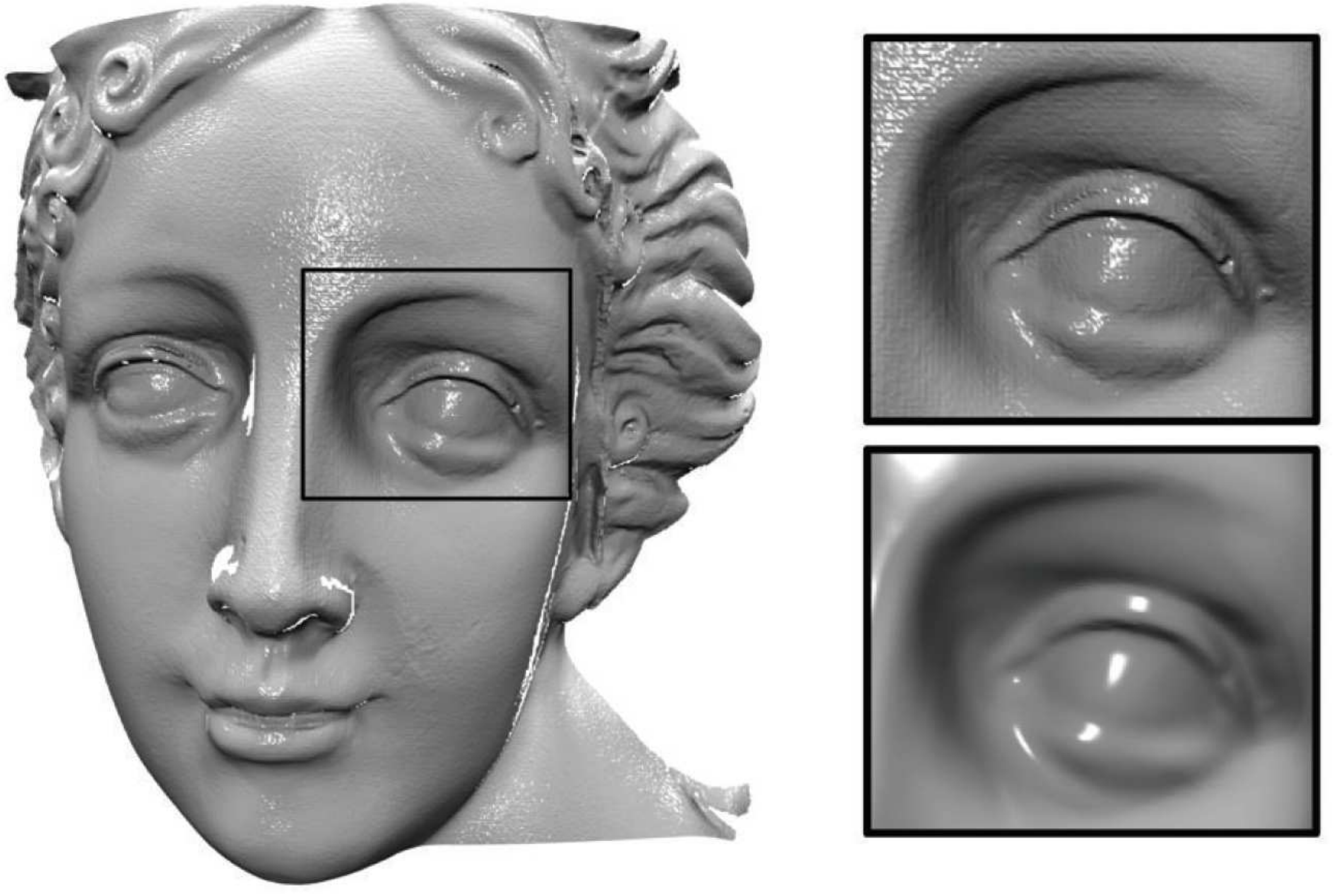}}
	\subfigure[Fairing with curvature energy minimization \cite{desbrun1999implicit}]{
		\label{fig:definition:fairing} 
		\includegraphics[width=1.65in]{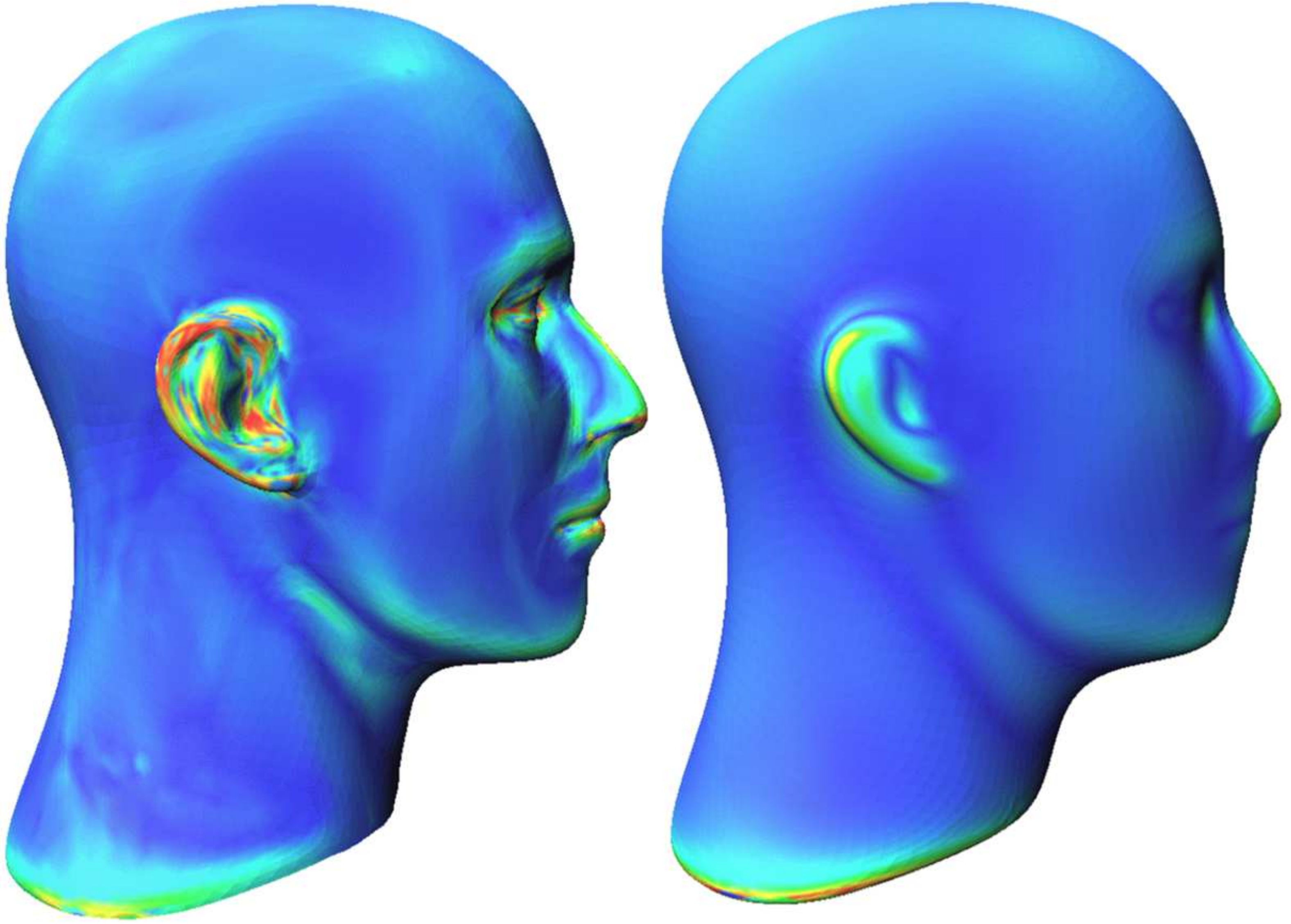}}
	\caption{A visual comparison of mesh denoising and fairing.}
	\label{fig:definition} 
\end{figure}

Firstly, we wish to clarify several synonymous concepts in the literature of mesh denoising. Mesh smoothing contains two main aspects: denoising and fairing \cite{botsch2010polygon}.
The goal of denoising is to generally remove certain high-frequency noise or information and preserve genuine information at all frequencies (see Fig. \ref{fig:definition:denoising}). In fairing, it not only smooths the surface, in order to remove the high-frequency noise, but also computes shapes that are as smooth as possible, by using constrained energy minimization (see Fig. \ref{fig:definition:fairing}). It is worth noting that surface fairing is used interchangeably with surface smoothing in many cases. In this work, we generally focus on the part of mesh denoising. 

The problem of mesh denoising has been extensively researched, and there has been a huge volume of mesh denoising approaches in recent decades. Among them, the traditional geometric methods often design customized object functions to smooth surfaces, based on certain prior assumptions. By contrast, learning-based techniques commonly focus more on an automatic and overall smoothing effect, regardless of the surface properties or noise distribution, with the aid of a large training dataset. 

Historically, mesh denoising has evolved in conjunction with the work of image filtering. For instance, bilateral mesh/normal filter evolved from gray and color image filtering \cite{TomasiM98}, sparsity priors were first developed to deal with the edge-preserving image smoothing problem \cite{XuLXJ11}, non-local low-rank techniques also have exhibited impressive results in image processing \cite{GuZZF14} before their extended applications in the 3D domain. Last but not least, advanced deep learning techniques in image domain, also undoubtedly inspire many 3D mesh surface denoising works.

Generally, we can place all the mesh denoising methods into a general framework, which can be split into several typical sub-steps. These sub-steps include coarse mesh denoising, vertex classification, normal estimation, and final vertex position updating. Any one of the existing methods covers one, two or full these sub-stages. For example, some one-stage methods \cite{taubin1995signal,desbrun1999implicit,vollmer1999improved,taubin2001linear,FleishmanDC03,Jones03,Wang06,ji2006non,he2013mesh,zhao2018robust} directly move the mesh vertices to their suitable positions. Although this kind of method is simple, they may either over-smooth sharp features, or cannot fully eliminate all noise. To alleviate the feature blurring problem, many researchers \cite{yagou2002mesh, shen2004fuzzy, lee2005feature,yoshizawa2006smoothing,sun2007fast,Zheng11,cgf/ZhangDZBL15,LiWLL17,ZhongXWLL18,CentinS18,YadavRP19} propose to first filter facet normals and then synchronize the vertex position with corresponding newly computed facet normals. The basic observation of these methods is that the normal variation is more sensitive than the vertex position variation in reflecting the surface variation. Moreover, several recent learning-based mesh denoising methods \cite{WangLT16,WangHWWXQ19,WeiGHXZKWQ19,zhao2019normalnet,li2020normalf} also consist of the two sub-steps above: facet normal learning and vertex position adjustment. Note that some multi-stage denoising schemes, containing extra operations like vertex classification \cite{HuangA08,wei2014bi,bian2011feature,lu2015robust,YadavRP18,fan2009robust}, or pre-filtering \cite{avron2010,bian2011feature,TsuchieH12,wang2012cascaded,zhu2013coarse,wei2014bi,LuCS17,lu2017efficient,yadav2017mesh,WeiHXLWQ19, LiLLWL19}, usually demonstrate more promising results, but with more time required. Besides, the intermediate result of each sub-step may also greatly affects the final result. In summary, every method contains one or more of these sub-steps. The key differences lie in how to solve these sub-steps effectively, efficiently and robustly, by exploiting various kinds of priors, assumptions, mathematical models and so on.

This paper surveys the advances in mesh denoising including both geometric methods and learning-based methods during the past decades. More detailed classification will be discussed in this work. We hope the survey provides readers with some insights into the state-of-the-art denoising methods and potential directions for future work. It may motivate researchers to develop new ideas for high-quality and robust mesh denoising. A benchmark is also built in this work, which includes detailed qualitative and quantitative evaluations of 16 advanced denoising algorithms. We will release all of these results in our specially-designed website.

The survey is organized as follows. In Section 2, we discuss several main challenges existing in mesh denoising. Section 3 provides two categorizations to give an overview of the existing denoising methods. From Section 4 to Section 6, three mainstream categories of denoising methods are discussed in detail. Section 7 concludes several mesh vertex position adjustment methods. Furthermore, a general comparison is demonstrated in Section 8 to give readers a basic evaluation of several state-of-the-art approaches. Conclusions and potential directions for future research are presented in Section 9.

\begin{figure}
	\centering
	\includegraphics[width=3.4in]{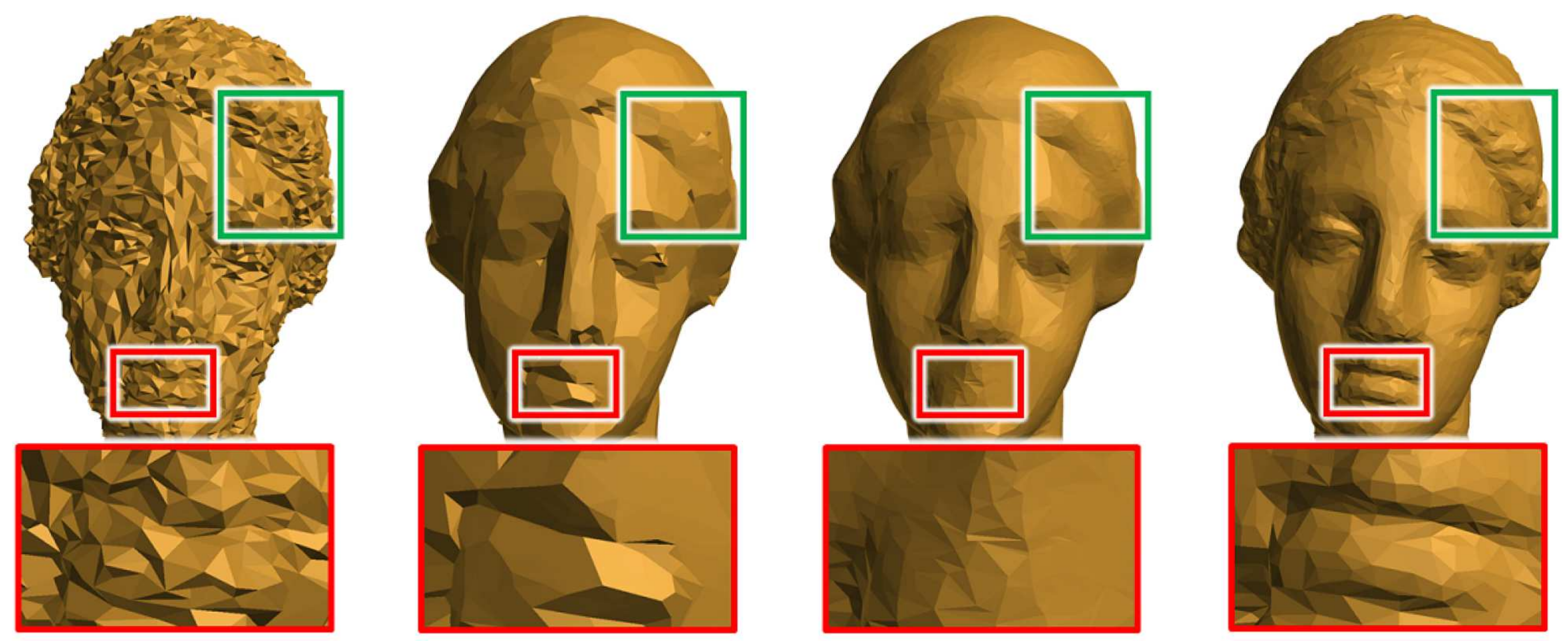}
	\caption{An example of mesh feature blurring and over-sharpening. From left to right: the noisy input model, the denoised results of \cite{he2013mesh} and \cite{WangLT16}, and the original model. We can observe small-scale features are obviously over-sharpened and over-smoothed, from the two smoothing results. Note that this figure is taken from \cite{LiZFH18}.}
	\label{fig:feature} 
\end{figure}

\section{Main challenges in mesh denoising}
When designing any denoising approach, certain types of fundamental problems must be taken into consideration. Whether these denoising methods can handle these problems well is essential to users. We summarize these core problems, in order to give the reader some basic knowledge.

\textbf{Feature degradation:} We consider the feature degradation problem from two aspects: feature blurring and feature over-sharpening. On the one hand, as observed from the third sub-figure of Fig. \ref{fig:feature}, because noise and certain-scale features are both regarded as high-frequency data, it is difficult to distinguish features, when a model is polluted by noise. Consequently, both of them may be regarded as noise and smoothed out. On the other hand, for preserving sharp features, some methods tend to over-sharpen the fine details in the models, as shown in the second sub-figure of Fig. \ref{fig:feature}.

\begin{figure}
	\centering
	\includegraphics[width=3.4in]{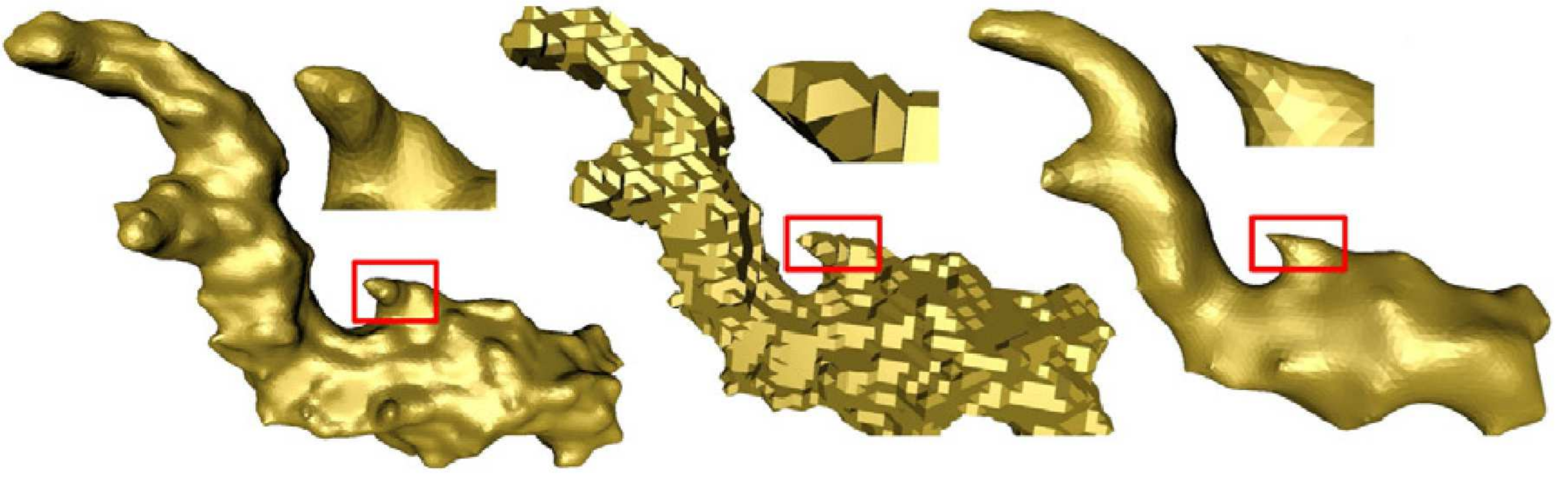}
	\caption{An example of model volume shrinkage. From left to right: the original model, the noisy input model, and a denoised result with heavy volume shrinkage. Note that this figure is taken from \cite{wei2018learning}.}
	\label{fig:volumn} 
\end{figure}

\textbf{Volume collapse:} Take a biomedical modeling case as an example, due to the parameter setting during CT/MRI scanning and organ segmentation, stepping noise and segmentation error often occur, which need to be eliminated. For this type of noise, isotropic filters achieve smoother results. However, it is very likely to cause serious deformation of the organ shape, as shown in Fig. \ref{fig:volumn}, which may affect the doctor's diagnosis.

\begin{figure}
	\centering
	\includegraphics[width=3.4in]{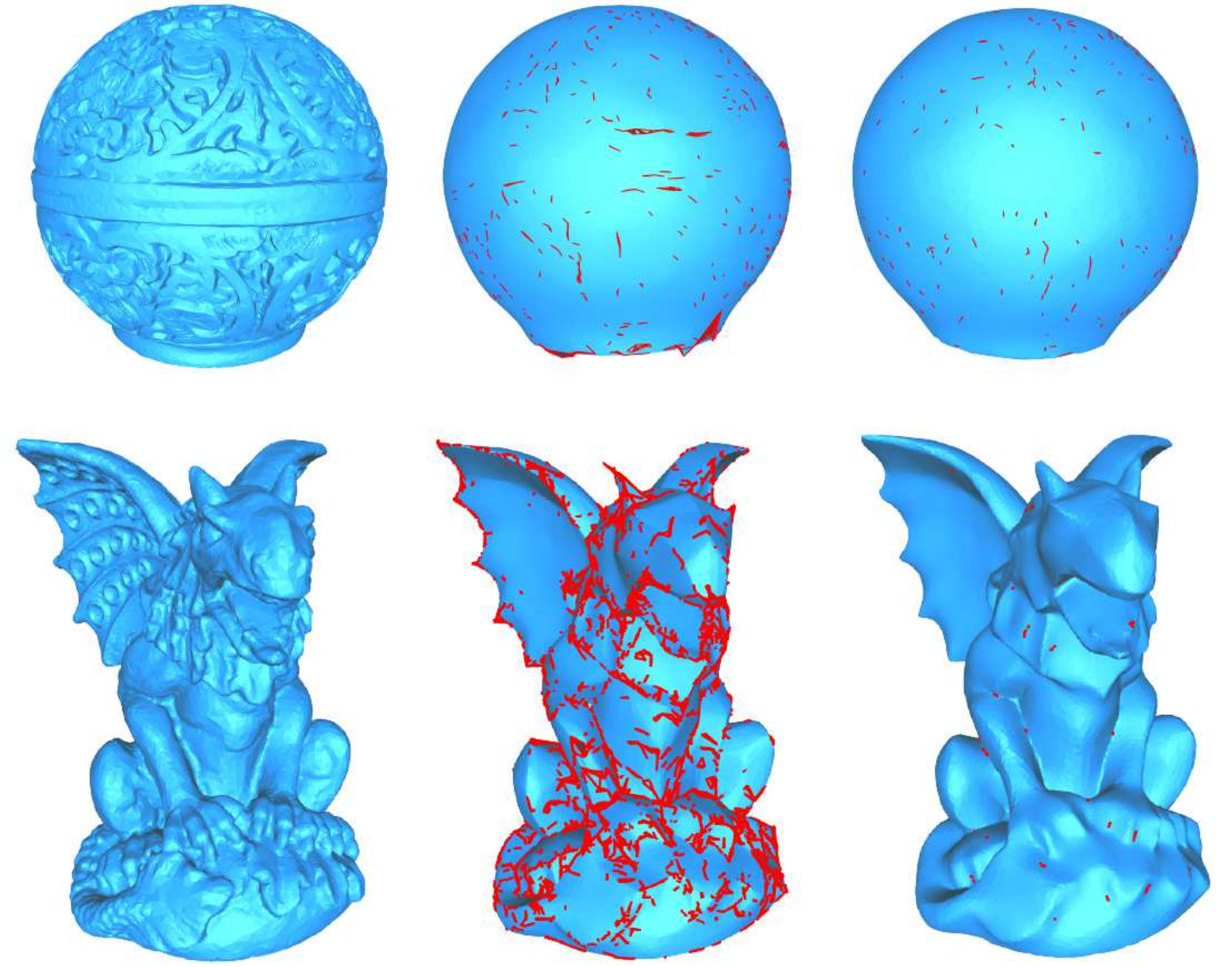}
	\caption{An example of mesh quality degradation. From left to right: the original model, the mesh updated by \cite{sun2007fast}, and the mesh updated by \cite{WangFLTLG15}. The flipped triangles are rendered in red. Note that this figure is taken from \cite{WangFLTLG15}.}
	\label{fig:degradation} 
\end{figure}

\textbf{Mesh facet quality degradation:} Generally speaking, the final step of denoising is always to move the vertex positions of the mesh, which is likely to cause triangle structure flipping, intersection and overlapping, as shown in Fig. \ref{fig:degradation}. 

\textbf{Hard to tune parameters:} Most mesh smoothing algorithms usually involve several key control parameters, which need to be adjusted empirically. This work is time-consuming and cumbersome. Moreover, for the same algorithm, even if the parameter values are slightly adjusted, the results may be significantly different.

\section{Classification of mesh denoising algorithms}
In this section, we suggest two different criteria for classifying the mesh denoising methods.

\subsection{Classification based on the ability of surface feature preservation}
An important principle of mesh denoising is the maintenance of geometric features. Most early works focused on isotropic algorithms that ignore sharp features (i.e., independent of surface geometry), while recent methods are anisotropic and attempt to preserve geometric features in the data. Therefore, from the perspective of feature preservation, we can classify existing methods into the following two categories.

\textbf{Isotropic methods:} Isotropic methods do not take the geometric features of the surface into consideration, and treat the features and noise equally as the high-frequency information. For example, Taubin et al. \cite{taubin1995signal} proposed a two-stage fairing method of surface signal processing, by alternately shrinking and expanding. Because of the linear complexity in both time and memory, this method can smooth large meshes very quickly. Later, Vollmer et al. \cite{vollmer1999improved} presented an improved Laplacian smoothing algorithm that filters noise efficiently but does not preserve features and shrinks the surface. Desbrun et al. \cite{desbrun1999implicit} introduced an implicit fairing method, by using implicit integration of a diffusion process that allows for efficiency, quality, and stability. Their method can also guarantee volume preservation during smoothing. Some other smoothing methods \cite{LiuBSP02,KimR05,NehabRDR05,SuWC09} are also considered as isotropic approaches, since they are all lack awareness of the surface features. Generally, isotropic denoising methods are effective to remove noise, but with the cost of features blurring, or even completely missing, and thus destroy desirable surface features. 

\textbf{Anisotropic methods:} Given that isotropic methods hardly preserve geometric features in the object, the focus of many recent works have been moved to anisotropic techniques  \cite{clarenz2000anisotropic,ohtake2000polyhedral,tasdizen2002geometric,FleishmanDC03,Jones03,hildebrandt2004anisotropic,bajaj2003anisotropic,sun2007fast,WangY11,WangY12,wang2012cascaded,GaoYH13,wei2013feature,zhu2013coarse,wei2014bi,yu2014feature,cgf/ZhangDZBL15,lu2017efficient,LuCS17,lu2015robust,WeiLPWLW17,LiZFH18,WangHWWXQ19,lu2018low}. Most of these methods are derived from the image processing domain, such as the anisotropic diffusion and classical bilateral filtering \cite{TomasiM98} in image processing. 

Anisotropic diffusion based methods \cite{ohtake2000polyhedral,ClarenzDR00,tasdizen2002geometric,BajajX03,hildebrandt2004anisotropic,ZhangH07, OuafdiZK08} take sharp features into account, such as the curvature tensor and the normal information of the mesh. For example, Ohtake et al. \cite{ohtake2000polyhedral} proposed an adaptive smoothing method, which incorporates the local curvature information and allows the reduction of possible oversmoothing. Tasdizen et al. \cite{tasdizen2002geometric} extended the anisotropic diffusion, a PDE-based, edge-preserving, image-smoothing technique, to surface processing. Their 3D smoothing version can remove complex, noisy surfaces, while preserving (and enhancing) sharp, geometric features. Hildebrandt et al. \cite{hildebrandt2004anisotropic} defined an anisotropic mean curvature vector, which was used for feature-preserving noise removal. 

Another kind of anisotropic denoising method has been to generalize the bilateral filter from the image domain to 3D geometry. Fleishman et al.~\cite{FleishmanDC03} and Jones et al.~\cite{Jones03} pioneered extending the image bilateral filter to mesh denoising, by directly adjusting vertex positions. Others\cite{LeeW05,sun2007fast,Zheng11,SolomonCBW14,cgf/ZhangDZBL15,WangFLTLG15} apply the bilateral filter to the surface normals instead, using various normal similarity functions.

\subsection{Classification based on mathematical models} 
According to the mathematical model used for denoising, the existing mesh denoising methods can be divided into three main classes: optimization-based methods, filtering-based methods, and data-driven methods. Because we will introduce these three types of methods in detail in three later sections, we will not elaborate on them in this subsection.

\textbf{Optimization-based method:} For this set of techniques, the triangular mesh denoising problem is transformed into a geometric optimization problem. That is to say, based on the input mesh information and a set of constraints defined by the real geometric information or the noise distribution prior, an energy equation can be formulated. Solving the optimization function is equivalent to the geometry smoothing process.

\textbf{Filter-based method:} The filter-based methods are typically extended from the image smoothing domain. Although the triangular mesh has the characteristics of arbitrary topology and sampling irregularity, by adjusting the parameters, the filter-based methods have achieved impressive results, even if the input is contaminated by different kinds of noise and different levels of noise. Hence, this kind of method is currently the most commonly-used one.

\textbf{Data-driven method:} The data-driven method builds a universal regression framework, by learning the mapping from noisy inputs to real surfaces. No user tuning is required during the online operation phase. It has achieved very satisfactory results in many testing models.

\begin{figure}
	\includegraphics[width=3.4in]{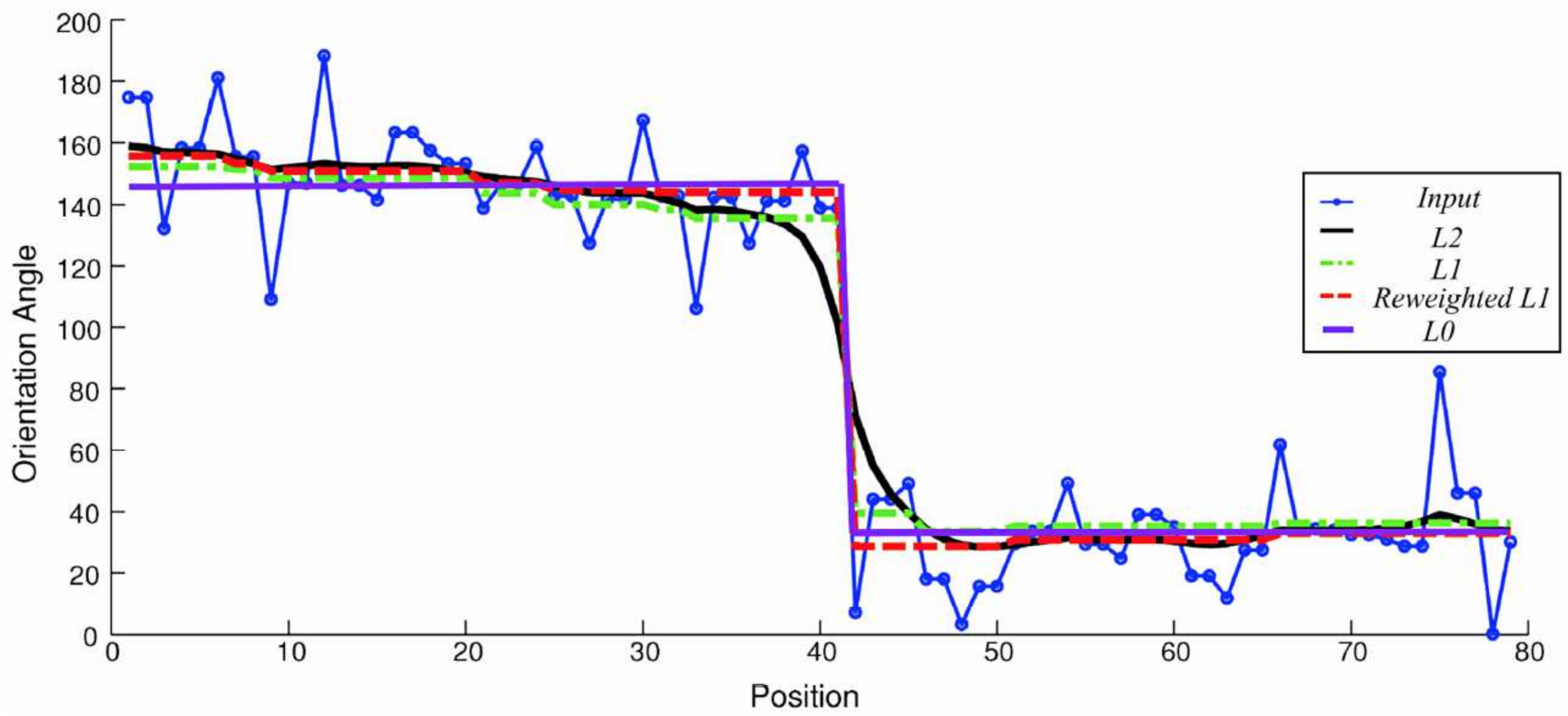}
	\caption{An illustration of smoothing a 1D signal by different norms. Note that this figure is taken from \cite{SunSW15}.}
	\label{Fig:sparsity}
\end{figure}

\section{Optimization-based methods} 
\subsection{Sparsity optimization} 
Sparsity-based optimization and sparse signal representation have proven to be an extremely powerful tool for processing signals, like audio, image, and video. Fig. \ref{Fig:sparsity} shows an illustration of smoothing a 1D signal by different error norms \cite{SunSW15}. In particular, in the image smoothing domain, some detailed components in the input image can be largely smoothed out with the sparsity regularization, while keeping large structures. Similarly, for the 3D mesh model, we can build many sparsity assumptions to remove noise.

He et al. \cite{he2013mesh} generalized the $L_0$ norm, which directly measures sparsity, from the image to the triangular mesh, to preserve prominent sharp features and smooth the remainder of the surface. In the $L_0$ minimization framework, the general optimization function can de formulated as follows:
\begin{equation}
\min _{\mathbf{s}} \left\|\mathbf{s}-\mathbf{s}^{*}\right\|^{2}_{2}+\lambda\|\nabla \mathbf{s}\|_{0},
\end{equation}
The first term keeps signals fidelity. The second one is to create a piecewise smooth signal, where $\nabla \mathbf{s}$ is a vector of gradients of the signals. As we can see from Fig. \ref{Fig:sparsity}, by minimizing the $L_0$ norm of the signal gradients, small signal fluctuations can be well smoothed, while preserving the most significant signal variation. Hence, it is intuitive to extend the $L_0$ norm to the 3D domain. For 3D mesh surfaces, a discrete differential operator, served as the function of gradient and describing the smoothness of local structures, is necessary. During the test, He et al. \cite{he2013mesh} found that using the discrete vertex-based Laplacian operator may fail to recover sharp features and shrink the surface away from the features. Therefore, they proposed to generalize the vertex-based cotan operator to an area-based edge operator. Notably, the key observation of $L_0$ minimization is that a noise-free mesh should be smooth, except for its geometric feature regions. Hence, it achieves better results for CAD-like input models with piece-wise smooth surfaces. However, when facing mixed noises or texture-rich models, this method may yield apparent visual artifacts, like sharp feature distortion or over-sharpening. In addition, the objective function of the $L_0$ smoothing algorithm is non-convex and nonlinear. Each iteration of the algorithm must solve a different large linear system, leading to relatively large time consumption. 

Since $L_0$ minimization is indeed a non-convex problem, He et al. \cite{he2013mesh} employed the same strategy in \cite{XuLXJ11}, by introducing auxiliary variables and alternating $L_0$ and $L_2$ minimization. However, $L_2$ term is sensitive to outliers and may propagate the errors to the $L_0$ term. Thus, the results calculated by their method are not sparse enough. To deal with this problem, Cheng et al. \cite{cheng2014feature} proposed a new approximation algorithm for the $L_0$ gradient minimization problem. This algorithm is based on a fused coordinate descent framework and is able to obtain a solution with good gradient sparsity and sufficiently close to the original input. Cheng et al. \cite{cheng2014feature} applied this scheme to smooth mesh facet normals and their method achieves a better sharp feature preservation effect.

Later, in the same framework of $L_{0}$ minimization, Zhao et al. \cite{zhao2018robust} presented a novel sparse regularization term to measure the sparsity of geometric features and distinguish features from noises. Both vertex positions and facet normals are optimized in a $L_0$ framework to faithfully remove noises and preserve features. Furthermore, an improved alternating optimization strategy was also developed in \cite{zhao2018robust} to address the $L_0$ minimization problem with guaranteed convergence and stability.

In view of the difficulty and long time required to solve the $L_0$ norm, \cite{WangYLDC14,WuZCF15,LuCS17} replaced the $L_0$ norm with the $L_1$ norm and turned to solve a convex optimization problem. In particular, based on the compressed sensing theory, Wang et al \cite{WangYLDC14} presented a $L_1$-analysis compressed sensing optimization to recover sharp features from the residual between the base mesh and the input mesh. The base mesh is first computed by a global Laplacian regularization denoising scheme. The main insights in this work lie in two aspects: 1) it has been proven that the unknown reasonable signals can be recovered, if they are sparse in the standard coordinate basis or sparse with respect to some orthogonal basis; 2) the $C^0$ sharp features on the shape can be sparsely represented in a coherent dictionary. Both theoretical and experimental results have shown that this method can faithfully decouple noise and sharp features.

Recently, under the observation that the Total Variation \cite{ZhangWZD15} and $L_0$ \cite{he2013mesh} based minimization methods can preserve sharp features well, but generate the side-effect of staircase artifacts, Liu et al. \cite{LiuLZW19} presented a high order normal filtering model with dynamic weights for preserving sharp features and removing the staircase effect in smooth regions simultaneously. The dynamic weights are applied in the proposed model to significantly improve effectiveness for preserving sharp features. As stated in this work, the dynamic weights can penalize smooth regions more than sharp features, which can be applied to achieve the lower-than-$L_1$-sparsity effect. During our test, this method achieves better smooth results than $L_0$ \cite{he2013mesh}, especially for the regions with stepping noise (see from evaluations section).

\begin{figure}
	\centering
	\includegraphics[width=3.4in]{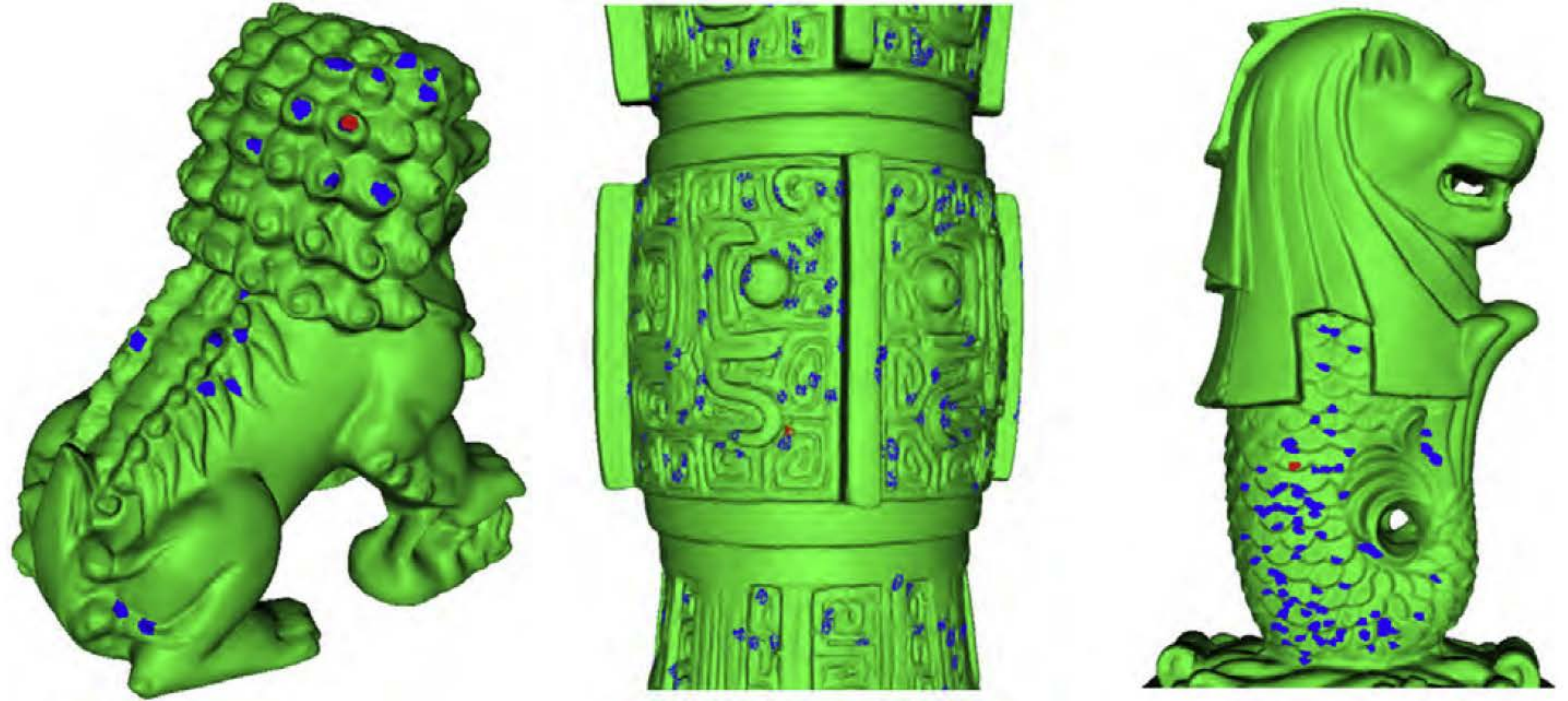}
	\caption{An illustration of non-local similarity. The red patch is the reference patch and the blue patches are similar patches. Note that this figure is taken from \cite{ChenHRXQGWW19}.}
	\label{fig:similarpatches} 
\end{figure}

\begin{figure}
	\centering
	\includegraphics[width=3.4in]{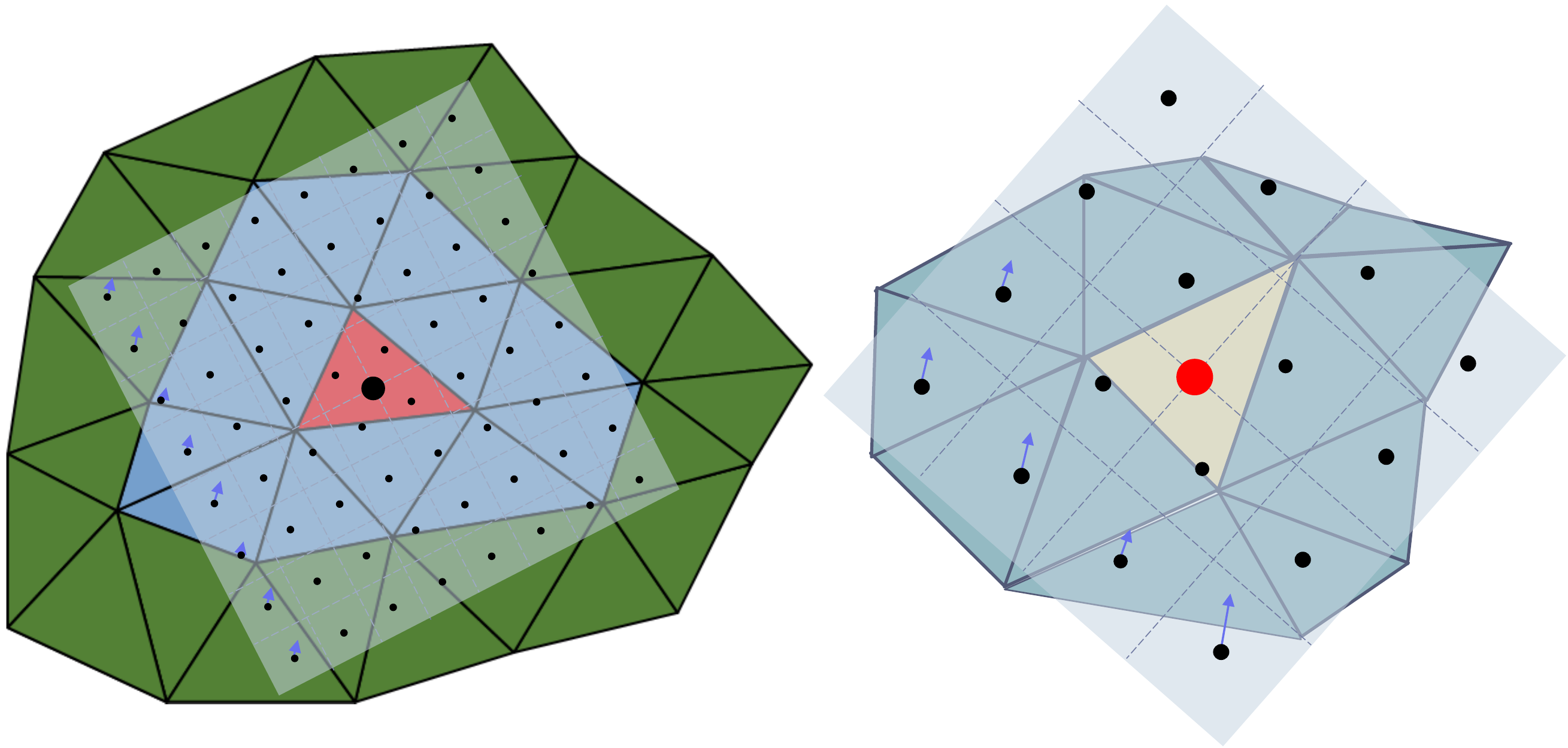}
	\caption{An illustration of patch regularity. Left: Sampling the bounding plane for a 2-ring local patch. Right: Sampling the bounding plane for a 1-ring local patch. The arrow means the projection from the sampled point to the original 3D patch. Note that this figure is taken from \cite{WeiHXLWQ19}.}
	\label{fig:2D} 
\end{figure}

\subsection{Low-rank optimization} 
In the last decade, the concept of non-local filtering has been extensively developed in various image processing applications, by combining the non-local self-similarity prior and the low-rank prior. It assumes that similar patches exist in the whole field of an image. If these similar patches are reshaped as vectors, they are highly linearly correlated. Thus, a patch matrix $\mathbf{G}$ constructed by stacking these patch vectors is low-rank. The general formulation as follows:
\begin{equation}
\mathop{\min}_{\mathbf{X}} \ \ rank(\mathbf{X}) + \lambda\|\mathbf{X} - \mathbf{G}\|_{F}^{2},
\end{equation}
where the first term is the rank of $\textbf{X}$, the second term represents the data fidelity measured by the square of the matrix $F$-norm, and $\lambda$ is a tradeoff parameter between the loss function and the low-rank regularization. 

Intuitively, this kind of method can also be extended to the 3D domain, due to the fact that the surface patches with the similar intrinsic properties always exist on the underlying surface of a noisy mesh (see from Fig. \ref{fig:similarpatches}). However, the methods developed in the image domain assume a 2D signal defined in a rectangular image domain with uniform grid sampling, so they cannot be used directly to handle 3D meshes. The main challenges are three-fold \cite{WeiHXLWQ19}: 

$\bullet$ Q1: How do you handle irregular connectivities and samplings of 3D surfaces?

$\bullet$ Q2: How do you define the similarity between any two 3D surface patches?

$\bullet$ Q3: How do you recover the low rank from a noisy matrix effectively?

Motivated by these above problems, some researchers have tried to solve them \cite{WeiHXLWQ19,LiZFH18,ChenHRXQGWW19,lu2018low}. They all follow a similar pipeline: local patch definition, similarity metric definition, similar patch identification, low-rank matrix construction, low-rank matrix recovery, and final surface calculation. Meanwhile, they all do not directly apply the low-rank recovery on the vertex position domain of the mesh, but on the normal field. We describe the main differences between the four methods, as shown in Tab. \ref{tab:low-rank steps} .

\begin{table*}[]
	\small
	\centering
	\setlength{\tabcolsep}{0.5mm}
	\caption{Comparison of the processing schemes for three challenges of the four low-rank mesh smoothing methods.}
	\label{tab:low-rank steps}
	\begin{tabular}{|c|c|c|c|c|}
		\hline
		\textbf{Methods} & \textbf{Q1} & \textbf{Q2} & \textbf{Q3} \\ \hline
		\textbf{\cite{WeiHXLWQ19}} & 2D parameterization  & normal voting tensor    & kernel low-rank recovery     \\
		\textbf{\cite{LiZFH18}}    & ring-based ordering scheme for local vertex patch  & guided normal patch covariance       & truncated $\gamma$ norm    \\
		\textbf{\cite{ChenHRXQGWW19}}  & corresponding nearest points within 2-ring neighborhood  & iterative closest point  & nuclear norm minimization   \\
		\textbf{\cite{lu2018low}}  & local isotropic structure  & tensor voting  & improved weighted nuclear norm minimization   \\
		\hline
	\end{tabular}
\end{table*}

\begin{figure}
	\centering
	\includegraphics[width=3.4in]{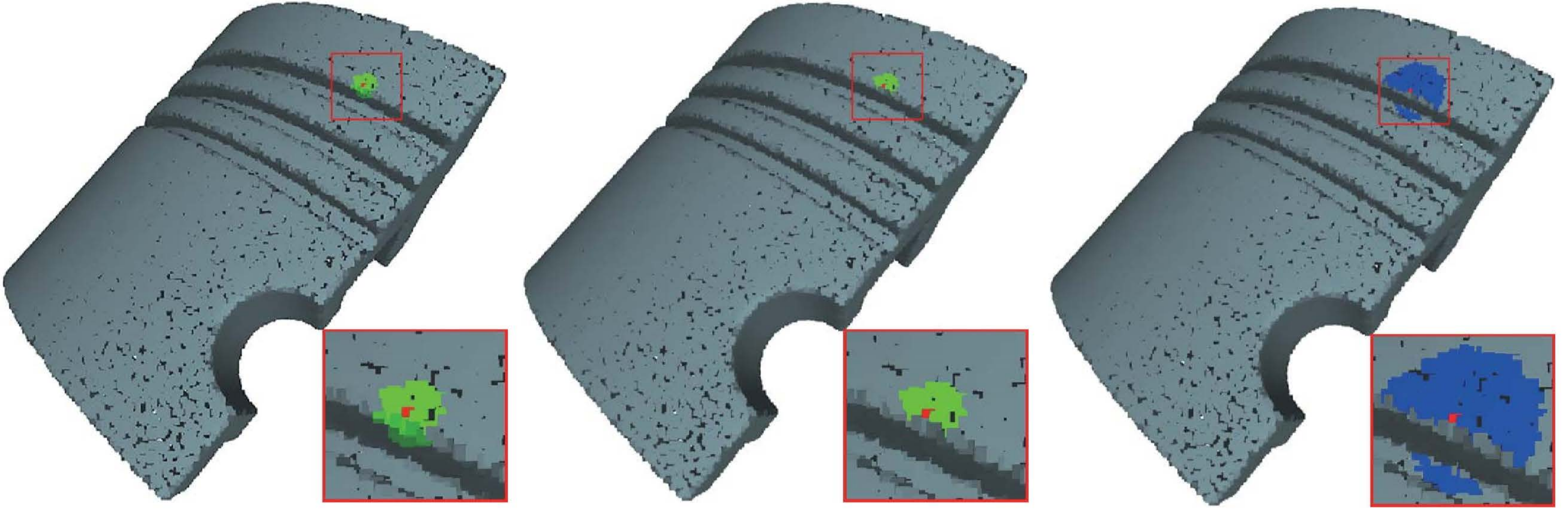}
	\caption{An illustration of similar non-local isotropic structures in \cite{lu2018low}. From left to right: the local structure (green points) of the centered red point, the local isotropic structure (green) of the red point, and the similar local isotropic structures of the local isotropic structure denoted by the red point. Each blue point denotes its isotropic structure. Note that this figure is taken from \cite{lu2018low}.}
	\label{fig:lu-lis} 
\end{figure}

\textbf{Wei et al.'s  method:} For each reference facet, Wei et al. \cite{WeiHXLWQ19} first collected a set of facets with the most similar local surface properties from the whole mesh, and constructed a local patch for each of these facets. To measure the similarity between any two local patches, they employed the robust normal tensor voting:
\begin{equation}
T_{f}=\sum_{f_{i} \in P(f)} \mu_{i} \mathbf{n}_{i} \mathbf{n}_{i}^{T},
\end{equation}
where $P(f)$ denotes the facet $f^{\prime}$s associated patch, $\mu_{i}$ is certain weight, and $\mathbf{n}_{i}$ is the normal of $f_i$. The similarity metric is defined as: 
\begin{equation}
\rho_{i, j}=\left\|\lambda_{1, i}-\lambda_{1, j}\right\|_{2}^{2}+\left\|\lambda_{2, i}-\lambda_{2, j}\right\|_{2}^{2}+\left\|\lambda_{3, i}-\lambda_{3, j}\right\|_{2}^{2},
\end{equation}
where $\lambda_{1} \geq \lambda_{2} \geq \lambda_{3} \geq 0$ are the eigenvalues of $T_{f}$.

To solve the problem of patch irregularity, they took a 2D parameterization scheme, as shown in Fig. \ref{fig:2D} . They then reshaped the normals of each regular patch as a patch vector and used them to construct a patch matrix $\mathbf{G}$. They further pursued low-rank matrix recovery in the kernel space for handling the nonlinear structure contained in the data, by making use of the half quadratic minimization and the specifics of a proximal based coordinate descent method. They finally obtained reliable normals from each recovered matrix, which are delivered to the bilateral normal filter as the guidance for further mesh denoising. 

\textbf{Li et al.'s method: } Different from Wei et al. \cite{WeiHXLWQ19} searching similar facets, Li et al. \cite{LiZFH18} searched for each vertex $\mathbf{v}_{i}$ on the input mesh a set of similar vertices. To determine the similarity between two vertices, they defined the normal patch covariance (NPC) descriptor to describe the geometry of local neighborhood around a given vertex:
\begin{equation}
\mathbb{C}\left(\mathbf{v}_{i}\right)=\frac{1}{N_{p}} \sum_{l=1}^{N_{p}}\left(\mathbf{n}_{l}-\overline{\mathbf{n}}_{i}\right)\left(\mathbf{n}_{l}-\overline{\mathbf{n}}_{i}\right)^{T},
\end{equation}
Actually, the above equation is a normal-based PCA. To effectively calculate the distance between two covariance matrices, this approach defined the NPC distance between vertices $\mathbf{v}_{i}$ and $\mathbf{v}_{j}$ as:
\begin{equation}
d_{\mathrm{NPC}}\left(\mathbf{v}_{i}, \mathbf{v}_{j}\right)=\sqrt{\left(\overline{\mathbf{n}}_{i}-\overline{\mathbf{n}}_{j}\right)\left(\mathbb{C}\left(\mathbf{v}_{i}\right)+\mathbb{C}\left(\mathbf{v}_{j}\right)\right)^{-1}\left(\overline{\mathbf{n}}_{i}-\overline{\mathbf{n}}_{j}\right)^{T}},
\end{equation}
Due to the noise impacts, the above metric computed from the noisy input is unstable. Hence, to further improve the selection of similar vertices, Li et al. took a pre-filtered mesh as a guidance signal to compute the guided normal patch covariance (G-NPC) distance:
\begin{equation}
d_{\mathrm{G}-\mathrm{NPC}}\left(\mathbf{v}_{i},\mathbf{v}_{j}\right)=d_{\mathrm{NPC}}\left(\mathbf{v}_{i}, \mathbf{v}_{j}\right) \cdot d_{\mathrm{G}}\left(\mathbf{v}_{i}, \mathbf{v}_{j}\right),
\end{equation}
where $ d_{\mathrm{G}}\left(\mathbf{v}_{i}, \mathbf{v}_{j}\right)$ is computed from the pre-filtered mesh.

This method then used the ring-based NPC ordering scheme to achieve regular normal patch structures, without using the 2D parameterization. Finally, an improved truncated $\gamma$ norm is utilized to recover the low-rank matrix.

\begin{figure}
	\small
	\centering
	\includegraphics[width=3.4in]{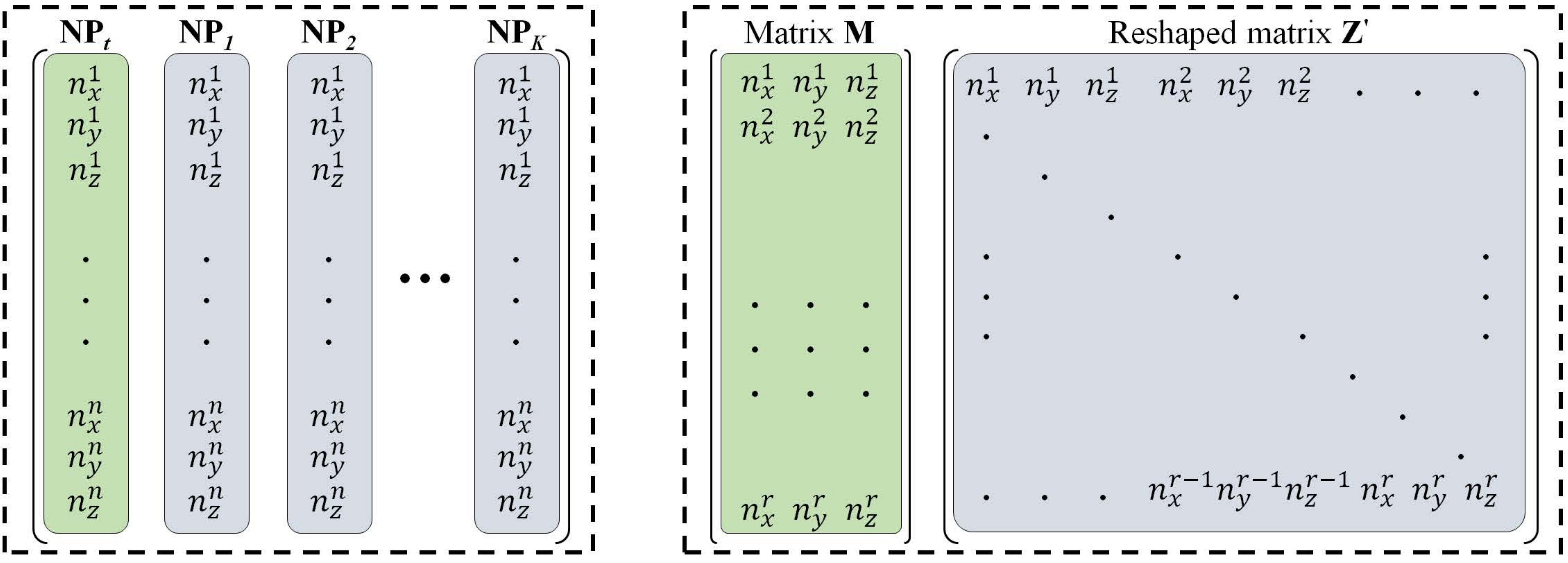}
	\caption{An illustration of non-local normal patch matrix constructed by \cite{WeiHXLWQ19,LiZFH18,ChenHRXQGWW19} (left) and \cite{lu2018low} (right).}
	\label{fig:npgm} 
\end{figure}

Followed the above two representative works, Chen et al. \cite{ChenHRXQGWW19} designed a joint low-rank recovery model to smooth out rich geometric features on a clean surface. At the same time, it is worth noting that Lu et al. \cite{lu2018low} proposed a new idea to utilize the low-rank prior for 3D geometry filtering. The interesting idea is that they filled the low-rank matrix with the normals from all similar local isotropic structures. That is to say, all the normals in their patch matrix are nearly the same. Fig. \ref{fig:npgm} shows an illustration of a normal patch matrix built by \cite{WeiHXLWQ19,LiZFH18,ChenHRXQGWW19} (left) and \cite{lu2018low} (right).

\section{Filter-based methods} 
Filter-based mesh methods usually assume that the surface noise is high-frequency, and denoise the noisy surface by a vertex-based or normal-based filter. The earlier methods belong to the vertex-based filtering method, for example, Laplace smoothing or its improved
versions \cite{taubin1995signal,desbrun1999implicit, pan2020hlo}. However, they will smooth out surface features. 

Later, more methods tend to perform filtering in the mesh normal field, since first-order normal variations better capture the local surface variations. Although there are various kinds of normal-based filters, all of them can be concluded in a general form:
\begin{equation}
\mathbf{n}_{i}^{t+1}= \Lambda (\sum_{j \in N(i)} w_{j} \mathbf{n}_{j}^{t})
\label{Eq:nf}
\end{equation}
where $\Lambda(\cdot)$ is a normalization operation, and $w_{j}$ is the weight that the neighboring normal $\mathbf{n}_{j}^{t}$ contributes to the updating of $\mathbf{n}_{i}^{t+1}$. The above function means the weighted average of the neighboring facet normals of the current facet. Actually, the key contribution of most normal-based filters is to compute the most reliable weight $w_{j}$. Specifically, the mean filter algorithm \cite{yagou2002mesh} defines the weight by the area of its neighboring facets. Following it, the alpha-trimming filtering approach proposed by \cite{YagouOB03} replaces the area-based weight by the normal differences between the current facet and its neighbors. The fuzzy vector median filtering \cite{shen2004fuzzy} first uses vector median filtering, and then uses Gaussian function to compute the weight, in order to obtain more accurate normals.

In addition to these above lateral (namely single weight) normal filters, another well-known filter is the bilateral (namely two weights) filter, which was first proposed by Tomasi et al. \cite{TomasiM98}. This kind of filter is able to preserve edges by means of a nonlinear combination of nearby image values. The bilateral filtering for image $I(\mathbf{u})$, at coordinate $\mathbf{u}= (x,y)$, is defined as:

\begin{equation}
\hat{I}(\mathbf{u})=\frac{\sum_{\mathbf{p} \in N(\mathbf{u})} W_{c}(\|\mathbf{p}-\mathbf{u}\|) W_{s}(|I(\mathbf{u})-I(\mathbf{p})|) I(\mathbf{p})}{\sum_{\mathbf{p} \in N(\mathbf{u})} W_{c}(\|\mathbf{p}-\mathbf{u}\|) W_{s}(|I(\mathbf{u})-I(\mathbf{p})|)},
\end{equation}
where $N(\mathbf{u})$ is the neighborhood of $\mathbf{u}$. This filter contains two monotonically decreasing Gaussian functions $ W_{c}(\cdot)$ and $ W_{s}(\cdot)$, which are used to measure the spatial weight and intensity weight, respectively. 

When extending this filter to the 3D domain, there are two main categories: filtering the vertex positions or facet normals. In the following, we will give the details on how to use the bilateral filter on mesh surfaces, as well as its improved variants.

\begin{figure}
	\centering
	\includegraphics[width=3.4in]{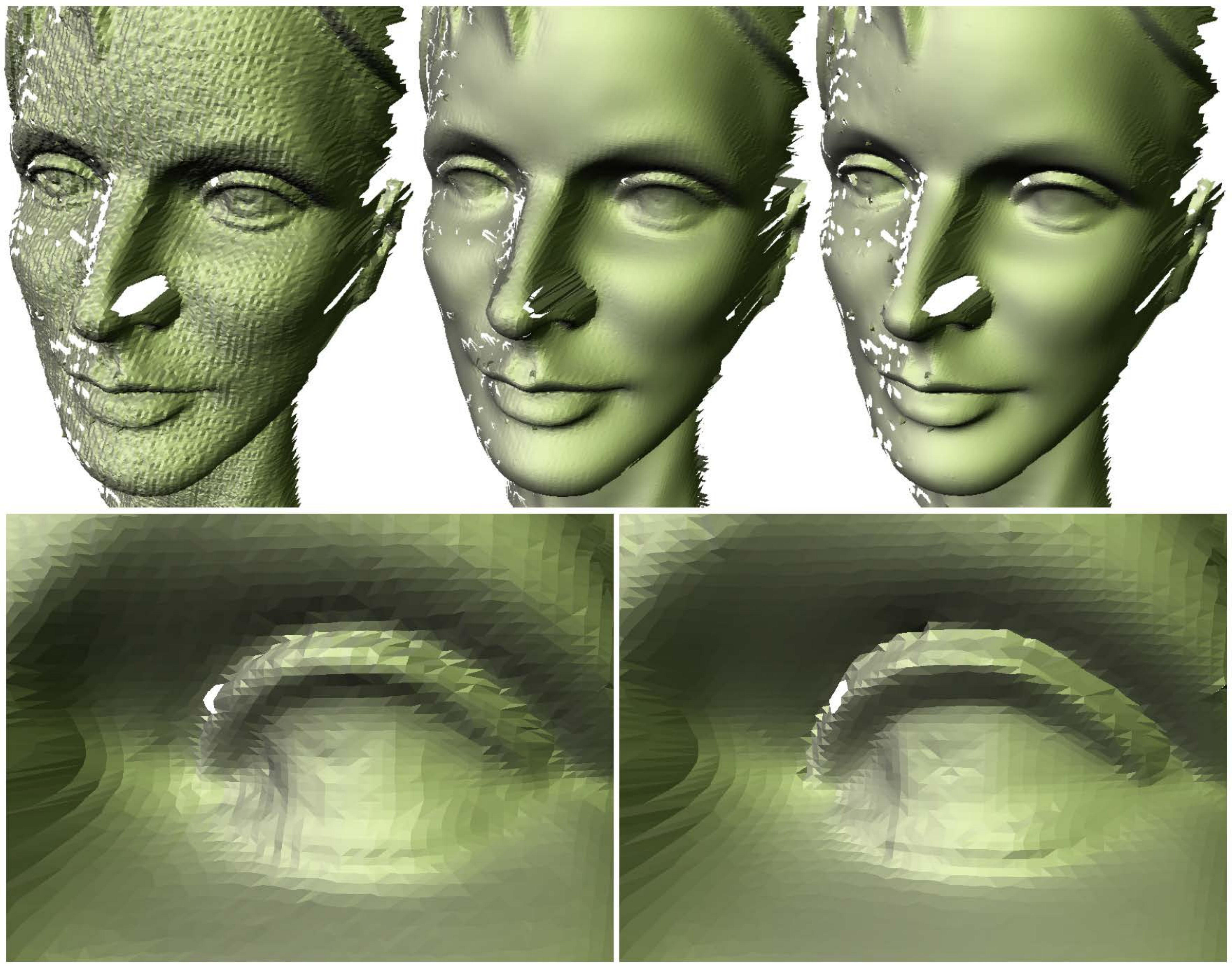}
	\caption{Results of denoising the face model. On the top row from left to right are the input noisy mode, the results of \cite{Jones03}, and \cite{FleishmanDC03}. On the bottom, the right eye of the model is zoomed, where the bottom left image shows the results of \cite{Jones03}, and on the bottom right is the result of \cite{FleishmanDC03}. Note that this figure is taken from \cite{FleishmanDC03}.}
	\label{fig:fleshman} 
\end{figure}

\subsection{Bilateral filter on mesh vertex positions} 
Fleishman et al. \cite{FleishmanDC03} proposed to directly modify vertex positions in the normal direction via a weighted average. The two weights are i) the spatial weight that measures the closeness from the neighbor vertex to the target vertex, and ii) the offset weight that measures the distance between the neighbor vertex to the tangent plane defined by the target vertex and its normal. Jones et al. \cite{Jones03} also followed a bilateral filtering framework. Besides the similar spatial weight, they used a new influence weight that depends on the distance between the prediction and the original position of the target vertex. Fig. \ref{fig:fleshman} shows the comparison results between \cite{FleishmanDC03} and \cite{Jones03}. As observed, both of the two methods can maintain the sharp features to a certain extent, but the result of \cite{Jones03} is less smoothed.

\subsection{Bilateral filter on mesh normals} 
\subsubsection{Bilateral facet normal filter} 
Zheng et al. \cite{Zheng11} developed two versions of bilateral filter in the mesh facet normal field. Both of the two denoising schemes regard the facet normals as a surface signal parameterized on an input mesh and formulates the influence of both spatial difference and signal difference into bilateral weighting. Specifically, the signal weight is defined as:
\begin{equation}
\mathrm{W}_{s}(\left\|\mathbf{n}_{i}-\mathbf{n}_{j}\right\|)=\exp \left(-\frac{\left\|\mathbf{n}_{i}-\mathbf{n}_{j}\right\|^{2}}{2 \sigma_{s}^{2}}\right)
\label{Eq:Ws}
\end{equation}
The spatial weight is defined as:
\begin{equation}
\mathrm{W}_{c}(\left\|\mathbf{c}_{i}-\mathbf{c}_{j}\right\|)=\exp \left(-\frac{\left\|\mathbf{c}_{i}-\mathbf{c}_{j}\right\|^{2}}{2 \sigma_{c}^{2}}\right)
\label{Eq:Wc}
\end{equation}
where $\sigma_{s}$ and $\sigma_{c}$ are both corresponding standard deviations, with which we can adjust the denoising and feature preservation power. Typically, $\sigma_{s}$ lies in the range of $[0.2, 0.6]$, and $\sigma_{c}$ is set as the average distance of all adjacent facets in an input mesh. 

The local version of mesh normal filtering is formulated as:
\begin{equation}
\mathbf{n}_{i}^{t+1}=K_{i} \sum_{j \in N(i)} \zeta_{i j} W_{c}\left(\left\|\mathbf{c}_{i}-\mathbf{c}_{j}\right\|\right) W_{s}\left(\left\|\mathbf{n}_{i}^{t}-\mathbf{n}_{j}^{t}\right\|\right) \mathbf{n}_{j}^{t}
\end{equation}
where $K_{i}=K(\mathbf{c}_{i})=1 / \sum_{j \in N(i)} \zeta_{i j} W_{c}(\|\mathbf{c}_{i}-\mathbf{c}_{j}\|) W_{s}(\|\mathbf{n}_{i}^{t}-\mathbf{n}_{j}^{t}\|)$ is  the normalization factor, $N(i)$ is the 1-ring facet neighborhood of a facet $f_i$, and $\zeta_{i j}$ is the weight to account for the influence from surface sampling rate. Notably, during the test, the authors found that the 1-ring facet neighborhood works well for non-CAD models and 2-ring is more suitable for CAD-like models. This is caused by more facets involved in the 2-ring neighborhood, and thus it is able to better characterize sharp features, especially sharp edges. 

\begin{figure}
	\centering
	\includegraphics[width=3.4in]{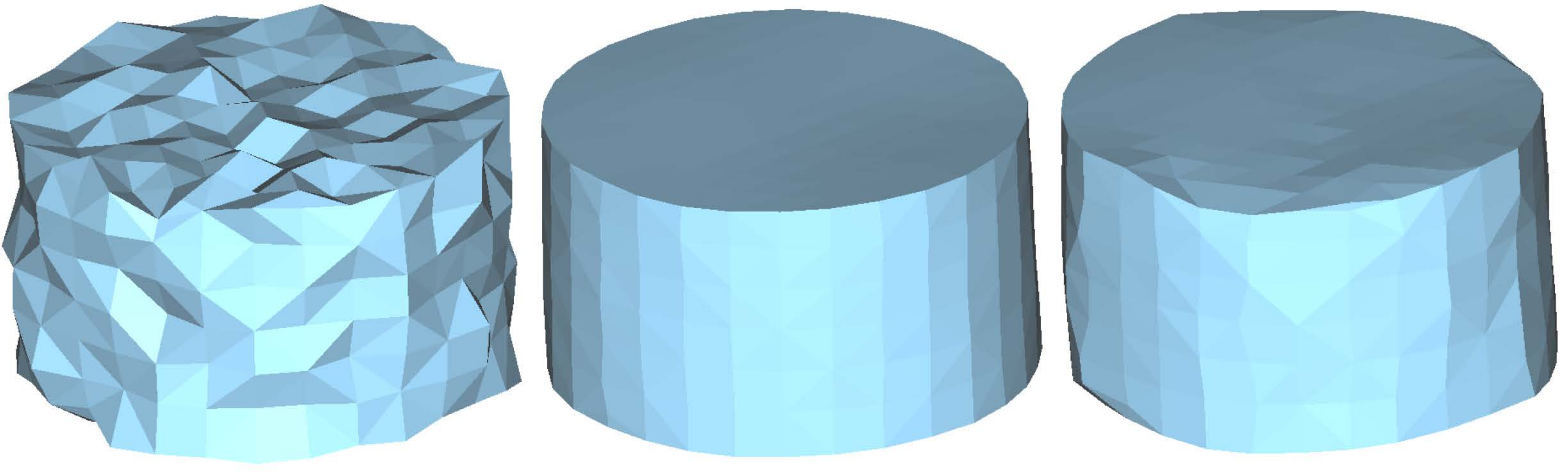}
	\caption{Denoising a noisy cylinder model (Left) by both local and global schemes in \cite{Zheng11}. The local scheme (Middle) better recovers geometric features than the global scheme (Right) when the level of noise is high. Note that this figure is taken from \cite{Zheng11}.}
	\label{fig:zheng} 
\end{figure}

Applying a local bilateral filter iteratively is able to propagate the filtering effect throughout the whole surface. Apart from it, Zheng et al. \cite{Zheng11} also proposed an alternative solution to update all the new normals in a single pass by minimizing:
\begin{equation}
E=\sum_{i} A_{i}\left\|\mathbf{n}_{i}^{\prime}-K_{i} \sum_{j \in N(i)} \omega_{ij} \mathbf{n}_{j}^{\prime}\right\|^{2} + \lambda \sum_{i} A_{i}\left\|\mathbf{n}_{i}^{\prime}-\mathbf{n}_{i}\right\|^{2}
\label{Eq:bnf}
\end{equation}
where $\mathbf{n}_{i}^{\prime}$ are the unknown normals for the denoised mesh, and $\omega_{i j} = \zeta_{i j}W_{c}W_{s}$ is the averaging weight in both Eq. \ref{Eq:Ws} and Eq. \ref{Eq:Wc} measured on the input mesh, $A_{i}$ is a weight of face area, and $\lambda$ is a balance parameter. The first term is the Laplacian normal smoothness term, and the latter one constrains the produced normals as similar to the original normals. By tuning $\lambda$, users can control the degree of denoising. In contrast, the local, iterative scheme is faster and has lower memory consumption than the global, non-iterative scheme. However, when the input is polluted by high-level noise, the local version produces more pleasing result (see from \ref{fig:zheng}).

\subsubsection{Guided normal filter} 
Although bilateral filter has achieved impressive performance, one obvious drawback is that the intensity difference weight is less reliable in the feature/structure regions. This is because the noisy input provides less reliable intensity (or the normal field in the 3D mesh surface) information. To resolve this problem, the joint bilateral filter was further proposed in the image processing domain, for processing flash/no-flash image pairs \cite{petschnigg2004digital} and \cite{EisemannD04}. The key idea is that the intensity weight is computed from another image, called the guided image or guidance, instead of the input image. Theoretically speaking, if we can provide more reliable information about the image structure, joint bilateral filtering can produce better results.

\begin{figure}
	\centering
	\includegraphics[width=3.4in]{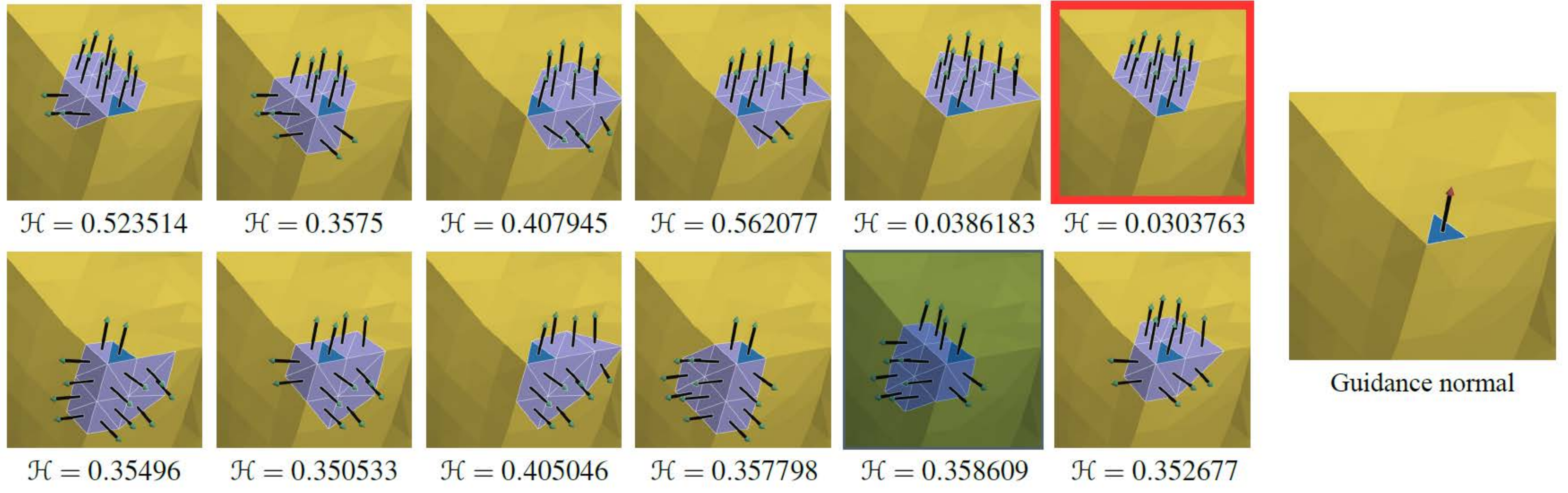}
	\caption{An illustration of the isotropic patch for computing the guidance normal. Note that this figure is taken from \cite{cgf/ZhangDZBL15}.}
	\label{fig:guidance} 
\end{figure}

Inspired from the success of the joint bilateral filter in image smoothing, Zhang et al. \cite{cgf/ZhangDZBL15} proposed a new triangular mesh normal filtering framework, called guided normal filter (GNF). Its formulation is written as:
\begin{equation}
\mathbf{n}_{i}^{t+1}=K_{i} \sum_{j \in N(i)} \zeta_{i j} W_{c}\left(\left\|\mathbf{c}_{i}-\mathbf{c}_{j}\right\|\right) W_{s}\left(\left\|\mathbf{gn}_{i}^{t}-\mathbf{gn}_{j}^{t}\right\|\right) \mathbf{n}_{j}^{t}
\label{Eq:gnf}
\end{equation}
where $\mathbf{n}_{i}^{t+1}$ is the filtered normal for facet $f_i$, and the $\mathbf{gn}_{j}^{t}$ is the so-called guidance. The main challenge lies in the construction of the guidance normal field, which needs to be defined in the same domain as the input, while providing enough information about the desired output. To realize it, Zhang et al. \cite{cgf/ZhangDZBL15} defined an isotropic neighboring patch for each facet. In this patch, the normals of all facets are nearly in the same direction (see from Fig. \ref{fig:guidance}). Then, the average normal of the chosen patch is used as the guidance normal for this facet. As we observed from Fig. \ref{fig:guidance}, the neighboring patch stays in the one side of a sharp feature and will not cross the edge, so the produced guided normal is feature-aware, and thus leads to better feature preservation results. 

A drawback of this approach is the strict requirement of correct guidance normal. This limitation was addressed in the work of \cite{ZhaoLWZ17,WangZLZL17,zhao2019graph}, but it is still hard to handle complex structures, such as narrow edges and corners.

\subsubsection{Rolling guidance normal filter} 
Instead of removing noise, smoothing geometric details (or geometry texture) like bumps, ridges, creases, and repeated patterns, is also essential for many applications \cite{WangFLTLG15}. Wang et al. \cite{WangFLTLG15} proposed a simple and effective scale-aware mesh smoothing filter, called the rolling guidance normal filter (RGNF), to process different scales of geometry features of triangular meshes. The key idea is to iteratively smooth out small geometric variations while preserving large-scale features, by applying an iterative joint bilateral filter to face normals. The iterative process can be summarized as:
\begin{equation}
\mathbf{fn}_{i}^{t+1}=K_{i} \sum_{j \in N(i)} \zeta_{i j} W_{c}\left(\left\|\mathbf{c}_{i}-\mathbf{c}_{j}\right\|\right) W_{s}\left(\left\|\mathbf{fn}_{i}^{t}-\mathbf{fn}_{j}^{t}\right\|\right) \mathbf{n}_{j}
\label{Eq:rgnf}
\end{equation}
where $\mathbf{fn}_{i}^{t+1}$ is the filtered normal for facet $f_i$. The main differences between RGNF and GNF lie in three aspects: 1) the first iteration of RGNF is the Gaussian filter in nature, since $\mathbf{fn}_{i}^{0} = \mathbf{0}$; 2) the filtered normal result $\mathbf{fn}_{i}^{t+1}$ is a weighted average of the original normal $\mathbf{n}_{j}$, which is constant during all iterations; 3) the guidance is the normals from last iteration. The principle of the rolling guidance filter can be understood as follows. In the first iteration, the filter is actually a Gaussian filter, thus the small geometric features whose scale is smaller than $\sigma_{s}$ can be smoothed effectively. Unfortunately, large structures are also smoothed to some degree. In the later iterations, $\mathbf{fn}_{i}^{t}$ is no longer zero. For the removed small features, their guidance normals are almost equal. So, the weight term $W_s$ approximately equals 1, and the filter is still a Gaussian filter, resulting in small structures still being removed. For large-scale structures, they are recovered gradually, since their guidance normals can provide effective weights, and thus joint bilateral filter sharpens the smoothed edges of large structures. It is worth noting that the task of geometry feature removal is different from denoising. As demonstrated in this work, it is clear that this method cannot recover the sharp feature well.

\subsubsection{Static/Dynamic filter} 
Recently, Zhang et al. \cite{ZhangDHPQL19} proposed a new approach for filtering signals defined on mesh surfaces, by utilizing both static and dynamic guidances. The static guidance means a guidance signal is obtained beforehand, and keeps unchanged during the filtering processing, while the dynamic guidance is just like the $\mathbf{fn}_{i}^{t+1}$ in RGNF, which is iteratively updated. Both of the two guidances have their own advantages and disadvantages: static guidance enables direct and intuitive control over the filtering process, but may be inconsistent with the input and lead to unsatisfactory results; dynamic guidance is adjusted according to the current signal values, but sensitive to noises and outliers. For taking benefits from both kinds of guidance, Zhang et al. \cite{ZhangDHPQL19} formulated the target function as:
\begin{equation}
\begin{aligned}
E_{\mathrm{SD}}=\sum_{i} A_{i}\left\|\mathbf{n}_{i}-\hat{\mathbf{n}}_{i}\right\|^{2} 
+ \lambda \sum_{i} \sum_{f_{j} \in N(i)}\left[A_{j} \cdot \phi_{\eta}\left(\left\|\mathbf{c}_{i}-\mathbf{c}_{j}\right\|\right)\right.\\\left.\cdot \phi_{\mu}\left(\left\|\mathbf{g}_{i}-\mathbf{g}_{j}\right\|\right) \cdot \psi_{\nu}\left(\left\|\mathbf{n}_{i}-\mathbf{n}_{j}\right\|\right)\right],
\label{eq:sd}
\end{aligned}
\end{equation}
The first term in the target function is a fidelity term that requires the output normal field to be close to the original normals, while the second term is a regularization for the output normal field. Specifically, $(\mathbf{g}_{i}, \mathbf{g}_{j})$ are from static guidance and $(\mathbf{n}_{i}, \mathbf{n}_{j})$ are dynamic current normals.
Extensive experimental cases show that this method is able to effectively smooth surface noise, or small-scale geometric features. 

\subsubsection{Bi-Normal filter} 
As we observed, the above mentioned filters only use the facet normal field. Nevertheless, there are two common normal fields of a mesh surface, i.e., the facet normal field and the vertex normal field. As known, each vertex's normal can be directly computed, by averaging its neighboring facet normals. As a result, there may exist redundant geometry information of the same mesh surface in the two fields. Hence, most of the normal based denoising methods can achieve quality results, by using only a single normal field \cite{wei2014bi}. However, Wei et al. \cite{wei2014bi} believed that these two types of normal fields also have some differences. Specifically, the facet normal field tends to reflect the global geometric variations of a mesh surface, while the vertex normal field is more illustrative of the local details of mesh vertices.

Based on this observation, Wei et al. \cite{wei2014bi} designed a novel strategy, called Bi-normal filtering, to more accurately compute the two normal fields and employ them to collaboratively guide the optimization of the vertex position for preserving features and dealing with irregular surface sampling. Firstly, all vertices in the input mesh are labeled as feature points and non-feature points, such that the mesh surface can be processed piecewise. The local bilateral normal filter is then used to initially estimate the facet normal field, for avoiding the influences of noise on the following steps. In order to more accurately estimate the vertex normal field and refine the facet normal field at the feature regions, a neighboring facets clustering scheme is proposed, which can provide an isotropic neighborhood for the normal estimation of feature vertices. Finally, all the mesh vertices are moved to new positions to simultaneously match the two normal fields.

\section{Data-driven methods} 
Data-driven methods have been widely used in image denoising. In the 3D domain, it has also been a research hotspot to develop a data-driven mesh denoising method. However, it is hard to directly apply learning-based techniques to irregular 3D mesh structures, in contrast to the regular grid structure of 2D images. To solve this problem, two intuitive ideas are: 1) designing a novel kind of network architecture, like PointNet \cite{qi2017pointnet} for unstructured point clouds; 2) extracting reasonable feature descriptors or representing the mesh surface in a regular format, so that many mature networks can be directly used. To the best our knowledge, although there are some new neural networks that are elaborately designed for mesh surface, none of them was used for mesh denoising. The recent several learning-based denoising work all belongs to the second type.

\subsection{Cascaded Normal Regression for mesh denoising}
Specifically, Wang et al. \cite{WangLT16} developed a filtered facet normal descriptor (FND) for expressing the geometry features around each facet on the noisy mesh and modeled a regression function with neural networks for mapping the FNDs to the facet normals of the denoised mesh. According to the filter used in constructing FNDs, two kinds of FND are designed:

1) Bilateral filtered facet normal descriptor (B-FND):
\begin{equation}
  \begin{split}
\mathbf{S}_{i}:= & (\mathbf{n}_{i}^{1}(\sigma_{s_{1}}, \sigma_{r_{1}}), \cdots, \mathbf{n}_{i}^{1}(\sigma_{s_{L}} , \sigma_{r_{L}}),\\
& \mathbf{n}_{i}^{2}(\sigma_{s_{1}}, \sigma_{r_{1}}), \cdots, \mathbf{n}_{i}^{2}(\sigma_{s_{L}}, \sigma_{r_{L}}),\\ 
& \cdots, \mathbf{n}_{i}^{K}(\sigma_{s_{1}}, \sigma_{r_{1}}), \cdots, \mathbf{n}_{i}^{K}(\sigma_{s_{L}}, \sigma_{r_{L}}))
  \end{split}
\label{eq:B-FND}
\end{equation}

2) Guided filtered facet normal descriptor (G-FND):
\begin{equation}
\begin{split}
\mathbf{S}_{i}^{g}:= & (\mathbf{gn}_{i}^{1}(\sigma_{s_{1}}, \sigma_{r_{1}}), \cdots, \mathbf{gn}_{i}^{1}(\sigma_{s_{L}}, \sigma_{r_{L}}), \\ 
&\mathbf{gn}_{i}^{2}(\sigma_{s_{1}}, \sigma_{r_{1}}), \cdots, \mathbf{gn}_{i}^{2}(\sigma_{s_{L}}, \sigma_{r_{L}}),\\ 
&\cdots, \mathbf{gn}_{i}^{K}(\sigma_{s_{1}}, \sigma_{r_{1}}), \cdots, \mathbf{gn}_{i}^{K}(\sigma_{s_{L}}, \sigma_{r_{L}})) 
\end{split}
\end{equation}
where each B-FND consists of a series of bilateral normal filtering \cite{Zheng11} results with different parameter pairs, including Gaussian parameters $(\sigma_{s_{i}}, \sigma_{r_{j}})$ and the max iteration number $K$, while each G-FND is computed by guided normal filtering \cite{cgf/ZhangDZBL15} with the same parameters. The key idea of this methods lies in the design of FND. In the past, when we use bilateral normal filter to remove noise, it is hard to find the most accurate parameters when the input mesh has features at different scales. Therefore, the denoised mesh is not always optimal. In this work, the designed FND models multi-scale features well and is robust to noise. Besides, with the help of the neural network, we do not need to do the tedious parameters tunning, because the network directly learns an optimal mapping from FND to the ground-truth normal results. Actually, we can intuitively and simply consider that the network learns a weighted combination of many bilateral normal filtering results, but best fitting the real accurate normal. Visual results and quantitative error analysis both demonstrated that this method outperforms the state-of-the-art mesh denoising methods and successfully removes different kinds of noise for meshes with various geometry features.

\subsection{Data-driven geometry-recovering mesh denoising}
As known, the magnitudes of geometric details are usually similar to and smaller than the noise’s, and thus noise and geometric details are commonly removed simultaneously by existing structure-preserving mesh filters. Moreover, existing denoising techniques have no mechanism to recover mesh geometry once it is lost. 

Motivated by this problem, and following the learning-based framework of \cite{WangLT16}, Wang et al. \cite{WangHWWXQ19} proposed to remove noise, while recovering compatible surface geometry, by a two-step learning scheme. Firstly, they learn the mapping function from the noisy model set to its ground-truth counterpart set using neural networks for removing noise, and then learn the reverse procedure of mesh filtering and recover geometry from denoised meshes using the learned regression function sequences. More simply put, there are two sub-networks: one is for removing noise, and another for recovering lost geometry. 

\textbf{Noise removal module: } In the first-step learning for noise removal, the B-FND (see Eq. \ref{eq:B-FND}) is directly used as the feature descriptor.

\textbf{Geometry recovery module: } In the second-step learning for geometry recovery, a new reverse B-FND is proposed. According to the image reverse filtering \cite{TaoZSWJ17}, the reverse normal filtering is formulated as:
\begin{equation}
\mathbf{n}^{k+1}=\Lambda\left(\mathbf{n}^{k}+\mathbf{n}^{*}-f\left(\mathbf{n}^{k}\right)\right)
\end{equation}
where $k$ denotes the iteration number, $\mathbf{n}^{*}$ is the initial filtered facet normal which is unchanged during the reverse filtering, and $f(\cdot)$ is the bilateral normal filtering operator. Then, the reverse B-FND (rB-FND) is defined as:
\begin{equation}
\mathbf{S}_{i_{2}}:=\left(\Lambda\left(2 \mathbf{n}_{i}-\mathbf{n}_{i}\left(\sigma_{s_{1}}, \sigma_{r_{1}}\right)\right), \ldots, \Lambda\left(2 \mathbf{n}_{i}-\mathbf{n}_{i}\left(\sigma_{s_{L}}, \sigma_{r_{L}}\right)\right)\right)
\end{equation}
In the training stage, Wang et al. \cite{WangHWWXQ19} mapped two large sets of B-FNDs and rB-FNDs to the corresponding facet normals of the ground-truth mesh, respectively. The extreme learning machine is employed as the regression method. Detailed quantitative and qualitative results on various data demonstrate that, this two-step learning algorithm competes favorably with the state-of-the-art methods, especially in terms of mesh geometry preservation.

\begin{figure}
	\centering
	\includegraphics[width=3.6in]{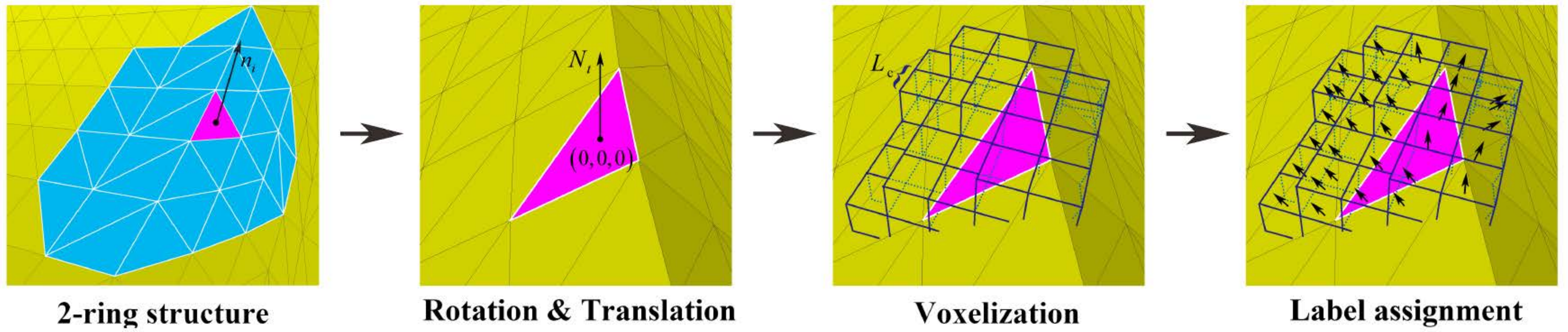}
	\caption{An illustration of voxelization strategy in \cite{zhao2019normalnet}. Note that this figure is taken from \cite{zhao2019normalnet}.}
	\label{fig:normalnet} 
\end{figure}

\subsection{Voxel-based NormalNet}
The voxelization strategy is another useful and intuitive scheme, by which we can convert the irregular local structure into the regular volumetric representation. Based on this representation, Zhao et al. \cite{zhao2019normalnet} designed a learning-based normal filtering scheme for mesh denoising, called NormalNet, to mimic the framework of GNF. Specifically, this scheme follows the iterative framework of filtering-based mesh denoising. During each iteration, firstly, the voxelization strategy is applied on each face in a mesh to transform the irregular local structure into the regular volumetric representation (see from Fig. \ref{fig:normalnet}). Therefore, both the structure and face normal information are preserved and the convolution operations in CNN can be easily performed. Secondly, instead of the RTV-based guidance normal generation and the guided filter in GNF \cite{cgf/ZhangDZBL15}, a deep CNN is utilized, which takes the ground-truth normal as the guidance, to simulate the filtering process of GNF. At last, the vertex positions are updated, according to the filtered normals. As known, the voxel-like structure consumes more memory and more training time, if we use very high voxel resolution. But low voxel resolution is hard to represent the shape information. Therefore, we think this is one drawback of this method.

\subsection{NormalF-Net: Normal Filtering Neural Network}
To avoid complicated voxelization operations for regularization, Li et al. \cite{li2020normalf} presented a novel mesh normal filtering network to mimic the low-rank recovery. This method combines non-local similarity (authors regard it as a geometry domain knowledge) and existing CNNs together. Specifically, this method follows the general pipeline of low-rank recovery based mesh denoising work. The biggest difference is that this method uses the CNN to mimic the function of low-rank recovery. Both objective and subjective evaluations demonstrate the superiority of this method in terms of noise removal and feature preservation. However, the non-local similar patches searching makes this method time-consuming, as we can see from the timing comparisons in \cite{li2020normalf}.

\subsection{Mesh Defiltering via Cascaded Geometry Recovery}
In this section, we introduce another nontraditional but practically meaningful work, mesh $defilter$ \cite{WeiGHXZKWQ19}. This reverse-filtering approach (termed as $DeFilter$) seeks to recover the geometry of a set of filtered meshes to their artifact-free status. This work is motivated by these observations: 1) there is huge volume of mesh denoising methods, thus resulting in many denoised results; 2) despite the great improvements, mesh denoising, however, is still an ill-posed problem, and no existing algorithm can serve as a mesh denoising panacea for various applications: they produce denoised results with a tradeoff between noise removal and geometry maintenance. 

Firstly, Wei et al. \cite{WeiGHXZKWQ19} smooth a set of ground-truth models to obtain their over-smoothed results. A set of mesh filters is also employed to generalize the over-smoothed models, which lose surface geometry to varying degrees. They then propose the generalized reverse Filtered Facet Normal Descriptor (grFND) as the feature descriptor and formulate each ground-truth facet normal as a function of the over-smoothed facet’s grFND. Next, they learn the function between the grFNDs and the ground-truth facet normals by the N-EML. Multiple iterations are required, since the first regression function coarsely finds the correspondence from the grFNDs to the ground-truth normals. Many results have verified the performance of the DeFilter thoroughly. It can recover the geometry of most denoised meshes without needing to know the exact filter used previously.

\section{Vertex updating} 
As we can see from Sections 4 to 6, we introduce many normal filtering techniques. However, we need to keep in mind that mesh smoothing seeks to adjust mesh vertex positions to the best estimates of their true positions. Hence, in these works, the next step after normal filtering is the vertex position updating. Moreover, in many cases, a single iteration cannot yield satisfactory results, and they often require multiple iterations. In the following, we give a brief introduction of several typical vertex position updating approaches.

\subsection{Locally iterative updating} 
Based on the observation that the filtered normal should be orthogonal to the three edges of each triangular facet in the mesh, Taubin \cite{taubin2001linear} formulated the vertex updating problem in a least-squares sense as follows: 
\begin{equation}
E(V)=\sum_{f \in F} \sum_{(i, j) \in \partial f}\left(\mathbf{n}_{f} \cdot\left(\mathbf{v}_{i}-\mathbf{v}_{j}\right)\right)^{2}
\end{equation}
where $\partial f$ denotes the set of edges that constitute the boundary of facet $f$. Classical gradient descent method can be directly used to minimize $E(V)$, so vertex position updating is implemented as: 
\begin{equation}
\mathbf{v}_{i}^{\prime}=\mathbf{v}_{i}+\lambda \sum_{j \in N_{v}(i)} \sum_{(i, j) \in \partial f_{k}} \mathbf{n}_{k}^{\prime}\left(\mathbf{n}_{k}^{\prime} \cdot\left(\mathbf{v}_{j}-\mathbf{v}_{i}\right)\right)
\label{Eq:vertex:Taubin2}
\end{equation}
 where $N_{v}(i)$ is denoted as the 1-ring vertex neighborhood of a vertex $v_i$ and $\lambda>0$ is the learning rate. 

Since the choice of $\lambda$ is vital to the final results, to avoid to carefully set the value of $\lambda$, Ohtake et al. \cite{OhtakeBB01} explicitly fixed $\lambda = \frac{1}{6 \sum_{k \in F_{V}(i)} A_{k}} $. So, the vertex updating is reformulated as:
\begin{equation}
\mathbf{v}_{i}^{\prime}=\mathbf{v}_{i}+
\frac{1}{3 \sum_{k \in F_{V}(i)} A_{k}} \sum_{j \in N_{V}(i)} \sum_{(i, j) \in \partial f_{k}} A_{k} \mathbf{n}_{k}^{\prime}\left(\mathbf{n}_{k}^{\prime}\cdot\left(\mathbf{v}_{j}-\mathbf{v}_{i}\right)\right)
\end{equation}
where $A_{k}$ is the area of the triangle $f_k$. Unlike Taubin's algorithm, this method does not have the problem of choosing a learning rate, but it is computationally more expensive,  since it needs to compute triangle areas.

Later, to speed up the vertex updating scheme of \cite{OhtakeBB01}, Sun et al. \cite{sun2007fast} replaced the area weight by 1:
\begin{equation}
\begin{array}{l}{\mathbf{v}_{i}^{\prime}=\mathbf{v}_{i}+}
{\frac{1}{3|F_{V}(i)|} \sum_{j \in N_{V}(i)} \sum_{(i, j) \in \partial f_{k}} \mathbf{n}_{k}^{\prime}\left(\mathbf{n}_{k}^{\prime}\cdot\left(\mathbf{v}_{j}-\mathbf{v}_{i}\right)\right)}\end{array}
\label{Eq:vertex:sxf}
\end{equation}
This simple modification gives the same weight for each adjacent facet, regardless of its area. This is due to the fact that if a facet has a large area, there is generally a large distance between the its vertices, and vertices with larger distances from the target vertex should have a weaker influence on the position updating.

\subsection{Poisson deformation} 
For some models with large-scale geometric features, the filtered triangle mesh facet normal may change greatly. If we directly employ the above mentioned updating algorithm, large vertex offsets will occur, thus causing triangle flipping. 

To circumvent this problem, Wang et al.\cite{WangFLTLG15} regarded this normal-guided vertex updating problem as a mesh deformation problem and adopted the Poisson Mesh Deformation \cite{yu2004mesh} as their solver. The optimization function is formulated as: 
\begin{equation}
E_{\text {Poisson}}=\sum_{f_{i}} A_{i} \sum_{j=1}^{3}\left\|\nabla g_{i, j}-\nabla \hat{f}_{i, j}\right\|_{2}^{2}
\end{equation}
where $\nabla g_{i, j}, j=1,2,3$ are the gradient vectors on the face $f_i$ with unknown vertex positions, and $\nabla \hat{f}_{i, j}, j=1,2,3$ are gradients of the piecewise-linear basis nodal function defined on the triangle $\bar{f}_i$. $\bar{f}_i$ is obtained by rotating the original triangle $f_i$, according to the rotation transformation defined by the original normal and filtered normal. Minimizing this energy function concerning vertex positions leads to the well-known Poisson equation, which can be solved efficiently. 

However, in some cases, the geometry texture is so complicated that a rapid change of face normals will still cause triangle flipping in minimizing $E_{\text {Poisson}}$. To ease this problem, a gradient smoothness regularization term is added:
\begin{equation}
E_{\text {smooth}}=\sum_{e_{i j}} l_{e_{i j}} \sum_{k=1}^{3}\left\|\nabla g_{i, k}-\mathbf{R}_{i j} \nabla g_{j, k}\right\|_{2}^{2}
\end{equation}
where $e_{i j}$ represents an edge adjacent to faces $f_i$ and $f_j$, $l_{e_{i j}}$ is the edge length, and $\mathbf{R}_{i j}$ is the rotation matrix that rotates triangle $f_j$ along $e_{i j}$ to be in the same plane of $f_i$. The added smoothness term penalizes the quick changes, thus reducing triangle flips.

The final energy function can be written as:
\begin{equation}
E_{update}=\frac{E_{\text {Poisson}}}{\bar{A}}+ \lambda \frac{E_{\text {smooth}}}{\bar{l}_{\mathrm{e}}}
\end{equation}
where average face area $\bar{A}$ and average edge length $\bar{l}_{\mathrm{e}}$ are used to normalize $E_{\text {Poisson}}$ and $E_{\text {smooth}}$ under an unform scale, and $\lambda$ balances two energy terms. Larger $\lambda$ value reduces more flipped elements, while increasing the approximation error of the target normals. It is set as 0.5 by default.

\subsection{Bi-Normal updating} 
There is also another kind of updating scheme, namely, incorporating both vertex normal and facet normal into the optimization function. The joint optimization problem proposed by Wei et al. \cite{wei2014bi} is formulated as follows.
\begin{equation}
\arg \min _{\mathbf{v}^{\prime}}\left[E_{1}\left(\mathbf{v}^{\prime}\right)+\lambda E_{2}\left(\mathbf{v}^{\prime}\right)\right]
\end{equation}

The first term is defined as:
\begin{equation}
E_{1}\left(\mathbf{v}^{\prime}\right)=\sum_{f \in N_{v}(f)}\left[\mathbf{n}_{f} \cdot \left(\mathbf{v}^{\prime}-\mathbf{c}_{f}\right)\right]^{2}
\end{equation}
where the first term encodes the sum of squared distances from vertex $\mathbf{v}^{\prime}$ to its one-ring facets, $\mathbf{n}_{f}$ is the filtered normal of facet $f$, and $\mathbf{c}_{f}$ is its triangle barycenter.

The latter term is written as:
\begin{equation}
E_{2}\left(\mathbf{v}^{\prime}\right)=\left(\mathbf{v}^{\prime}-\mathbf{v}\right) \cdot \left(\mathbf{v}^{\prime}-\mathbf{v}\right)-\left[\mathbf{n}_{v} \cdot \left(\mathbf{v}^{\prime}-\mathbf{v}\right)\right] \cdot \left[\mathbf{n}_{v} \cdot \left(\mathbf{v}^{\prime}-\mathbf{v}\right)\right]
\end{equation}
where $\mathbf{n}_{v}$ represents the unit normal of vertex $\mathbf{v}$. The term encourages the vertex to move along its normal direction in the optimization. Only minimizing the first term derived from filtered facet normals, inevitably leads to feature blurring and vertex drifts to some extent. Thanks to the second quadric energy term, using the estimated vertex normals can produce better results.

\section{Evaluations}
Given the wide diversity in mesh denoising methods, in this section, we first give a comprehensive comparison that describes the general abilities of the state-of-the-art methods dealing with different denoising challenges. Both visual and quantitative denoising results by different methods are then reported. To achieve a fair comparison, all of the results are provided by the authors, or produced by the open-source codes with our careful tunning. We summarize the parameters of these methods in Tab. \ref{tab:comp:parameter}. For some results provided by the authors, if without extra providing corresponding parameters, we directly leave it empty. The selected methods for the comparison cover all the common denoising categories. 

The methods involved are: Fleishman et al. \cite{FleishmanDC03} (BMF), Zheng et al. \cite{Zheng11} (BNF), Zhang et al. \cite{cgf/ZhangDZBL15} (GNF), Zhao et al. \cite{zhao2019graph} (GGNF), Lu et al. \cite{lu2015robust}, He et al. \cite{he2013mesh} ($L_0$), Li et al. \cite{LiZFH18} (NLLR), Liu et al. \cite{LiuLZW19} (HOF), Yadav et al. \cite{YadavRP19} (ROFI), Pan et al. \cite{pan2020hlo} (HLO), Zhang et al. \cite{ZhangDHPQL19} (SDF), Wang et al. \cite{WangLT16} (CNR), Zhao et al. \cite{zhao2019normalnet} (NormalNet), Wei et al. \cite{WeiHXLWQ19} (PCF), Wang et al. \cite{WangHWWXQ19} (DGRMD) and Li et al. \cite{li2020normalf} (NormalF-Net). Note that for some testing models, we only show partial denoised results, since we can not collect all the denoised results of these methods mentioned above.

All of the compared denoising results and the corresponding quantitative statistics will be included in our benchmark and released in our website. Future researchers can directly download them for a fast and fair comparison. We sincerely hope that other researchers could upload their corresponding results to this website, to further facilitate subsequent researchers. 

\subsection{Abilities for different challenges}
The main abilities of all denoising methods are concluded as the following four aspects:

$\bullet$ \textbf{A1}: Does it possess the capability of shallow/fine feature preservation? Note that for effectively removing noise, several existing methods may over-smooth the input surface, which causes the shallow/fine feature disappearing.

$\bullet$ \textbf{A2}: Does it possess the capability of avoiding feature over-sharpening? Note that for preserving sharp features, several existing methods may introduce a common extra artifact, namely feature over-sharpening.

$\bullet$ \textbf{A3}: Does it possess the capability of avoiding volume shrinkage?

$\bullet$ \textbf{A4}: Does it possess the capability of avoiding mesh quality degradation (triangle flipping, intersection and overlapping)?

From these above four aspects, Tab. \ref{tab:comp:description} reports an overall comparison for different kinds of methods, including the optimization-based approach (I), Filter-based one-stage approach (II), Filter-based two-stage approach (III), Filter-based multi-stage approach (IV), and Data-driven-based approach (V). We summarize these capabilities through experimental observation and theoretical analysis. To avoid the ambiguity in the evaluation, the symbol of ~\xmark~ represents that a certain method has a poor performance in this aspect, or even does not have this capacity, while the symbol of ~\checkmark~ on the contrary. The most important characteristic of each method is also summarized in the last column of Tab. \ref{tab:comp:description}. From this table, we can conclude that: 1) Filter-based two/multi-stage methods are able to preserve features well, since they often first compute a reliable mesh normal field; 2) Optimized-based methods do well in removing noise, but with the cost of smoothing fine details and high time expense; 3) the recent popular learning-based approaches achieve the most impressive performance, based on their capability in all the four aspects. 

\subsection{Visual Comparison}
The subjective comparison of five single models and three scenes with multiple objects is visualized in Figs. \ref{block}  , \ref{child} , \ref{fandisk} , \ref{boy01} , \ref{cone04}, and three additional figures in the supplemental material. The testing models contain: one CAD model with sharp features, one model with rich details, one model containing both sharp and shallow features, two raw scans acquired by Kinect from \cite{WangLT16} and three scenes scanned by a portable laser scanner. Each scene contains several CAD models with different levels of sharp features. We also add extra Gaussian noise on the latter two scenes.

Taking the Block model as an example (see from Fig. \ref{block} ), we find that BMF \cite{FleishmanDC03}, BNF \cite{Zheng11}, NLLR \cite{LiZFH18} and HOF \cite{LiuLZW19} achieve less pleasing results in some irregularly-sampled regions, due to lack of the mechanism for handling the problem of the sampling irregularity. 

For the model with rich details (see from Fig. \ref{child}),  we can observe that both $L_0$ \cite{he2013mesh} and SDF \cite{ZhangDHPQL19} over-smooth small-scale geometric features (the hair region), since $L_0$ pursues the feature sparsity in the whole surface, and SDF is designed especially for the mesh feature removal.

As observed in Fig. \ref{fandisk} , the shallow features can be recovered by most filter-based methods and all the learning-based approaches. The result produced by DGRMD \cite{WangHWWXQ19} is more satisfactory than that by CNR \cite{WangLT16}, because there is an elaborately designed mechanism in DGRMD that is designed to recover lost geometric features, when smoothing the input surface.

It is rather interesting to find that, from Figs. \ref{boy01} and \ref{cone04} , both the results produced by filter-based methods and those by learning-based methods are less satisfactory. One main reason is that the inputs acquired by Kinect are corrupted by heavy stepping noise, whose structure is similar to certain-scale sharp features. Therefore, these methods inspired by the bilateral filter may retain these fake features. For the learning-based methods, like CNR \cite{WangLT16} and DGRMD \cite{WangHWWXQ19}, their input descriptor is designed by bilateral filter, so their denoising performance is significantly limited by the effect of bilateral filter. NormalNet \cite{zhao2019normalnet} mimics the process of GNF \cite{cgf/ZhangDZBL15}, which still is a variant of bilateral filter. We can observe that NormalNet can produce a pleasing result better than GNF, but with residual noise. We also observe $L_0$ \cite{he2013mesh} cannot handle this kind of noise, because it also regards certain noise as geometric features. Among all of denoising results in Fig. \ref{cone04}, it is obvious that the surfaces of the results produced by HLO \cite{pan2020hlo} and SDF \cite{ZhangDHPQL19} are very smooth, due to the strong filtering capability of the two methods. However, we can find the expense of strong filtering is the the surface deviation, as illustrated in the Tab. \ref{tab:comp:error} (see from the metric value of $d$ of the Cone model).

We often measure a complex scene with multiple objects. To evaluate existing denoising methods on this type of data, we scan three scenes, each of which contains several CAD-like objects. We briefly denote the three scenes as Scene-1, Scene-2, and Scene-3. The detailed results are shown in the supplemental material. In the case of Scene-1, the noise-level is slight. We observe that most existing methods can produce satisfactory results. Especially, SDF \cite{ZhangDHPQL19} and $L_0$ \cite{he2013mesh} generate the flattest surfaces, while keeping sharp features well preserved. For the latter two scenes (Scene-2 and Scene-3), we add extra Gaussian noise. Generally, GNF \cite{cgf/ZhangDZBL15} produces better results than other filter-based methods, like BMF \cite{FleishmanDC03}, BNF \cite{Zheng11}, and ROFI \cite{YadavRP19}. SDF \cite{ZhangDHPQL19} preserves large sharp features rather well, but may over-smooth shallow geometric features. Among the learning-based approaches, the denoising result of NormalF-Net \cite{li2020normalf} still contains slight noise, while the latter two are more pleasing. We think the slight noise is generated, due to the fact that it is hard to keep the searched local patches sufficiently similar. 

\subsection{Quantitative Comparison}
Apart from the subjective comparison, quantitative statistics about denoising errors are also demonstrated. To evaluate the quality of the denoised surface, we introduce three classical metrics: 

1) The average angular difference (denoted as $\theta$), which is the mean angular difference of facet normals between the ground truth and the denoised mesh. A lower $\theta$ indicates a better result;

2) The root mean square of distance from the denoised mesh to the known ground-truth mesh (denoted as $d$). 

3) The average Hausdorff distance between the denoised mesh and the known ground-truth mesh (denoted as $hd$). It is normalized by the diagonal length of the bounding box of the mesh. 

Note that we do not use the metric $hd$, since some scanned ground-truth models often contain extra parts, which will bring error to the Hausdorff distance. We compare the objective performance on all of the testing models or scenes. The quantitative results of $\theta$ and $d$ are shown in Tab. \ref{tab:comp:error} . The best results are in bold. As shown in this table, the results produced by the learning-based techniques generally are more pleasing than traditional geometric methods. It is worth noting that the results yielded by PCF \cite{WeiHXLWQ19} have lower errors than many other methods. This is because PCF \cite{WeiHXLWQ19} pre-filters the mesh facet normal field first, via a non-local low-rank recovery scheme, and then the generated reliable normal field is fed to the GNF as an accurate guidance.  

Notably, we do not report a detailed running time statistics in this work, since many results are directly provided by authors. If possible, we will try our best to report the detailed times and running platforms in our website. 

\section{Conclusion}
The area of mesh denoising has grown from methods that design various kinds of geometric priors, to approaches that pursue a general and automatic solution. Our survey provides insight into this wide array of methods, highlighting the strengths and limitations that currently exist in the field. A brief discussion of future research directions are presented as follows: 

(1) In practice, we often need to scan large-size objects, thus obtaining real-scanned models with extremely large data size. In our testing, we find many existing methods either cannot process this kind of mesh, or may consume a lot of time. Especially, when the input is corrupted by heavy noise, we need more iterations or neighborhood scales.

(2) Although the deep learning techniques are rather useful in the image domain, their extension for 3D mesh processing still needs to be improved.  Moreover, as we investigated, very few learning methods exist for mesh denoising, while this is a very hot topic for point cloud processing \cite{RoveriOPG18, YuLFCH18PU, YuLFCH18, Casajus0R19, RakotosaonaBGMO20}.

(3) Considering the complexity of 3D structures, it is hard to directly apply mature 2D learning network architecture to mesh surface. The recent popular graph-learning technology may be a suitable network for mesh processing, since we can regard the mesh surface as a graph structure.

(4) Mesh denoising has actually been well-developed for many years. We think it will be meaningful to combine these traditional wisdoms with powerful machine learning technology together, to collaboratively solve the mesh denoising task.

(5) Lastly, there are relatively few real-scanned mesh datasets containing both ground-truth parts and noisy parts, which can be used for learning-based mesh denoising at present. This requires everyone's efforts to construct a dataset covering different kinds of models.


\label{main}

\ifCLASSOPTIONcaptionsoff
  \newpage
\fi



%

\bibliographystyle{IEEEtran}
\bibliography{main}

\begin{table*}[]
	\centering
	\footnotesize
	\setlength{\tabcolsep}{0.5mm}
	\caption{Parameters setting.}
	\label{tab:comp:parameter}
	\begin{tabular}{|c|c|c|c|c|c|c|c|c|}
		\hline
		\textbf{Methods} & Block  & Child & Fandisk & Boy & Cone & Scene-1  & Scene-2 & Scene-3\\ 
		\cline{1-9}
		$L_0$ \cite{he2013mesh}          & ($\lambda=0.6$, default) & (Default)& ($\lambda=1$, default) & ($\lambda=0.4$, default) & ($\lambda=1$, default) & ($\lambda=1$, default) & ($\lambda=1$, default) & ($\lambda=1$, default)\\
		HOF \cite{LiuLZW19}              & (5, 1) & (300, 1.1) & (50, 1.2) & (1800, 1) & (600, 1) & $--$ & $--$ & $--$\\
		HLO \cite{pan2020hlo}            & (5) & (5) & (4) & (20) & (50) & 5 & 10 & 10  \\
		NLLR \cite{LiZFH18}              & (0.6,10,10) &  (0.39,10,10) & (0.49,10,20) & (0.78,30,20) & (1,30,60) & $--$  & $--$ & $--$\\
		PCF  \cite{WeiHXLWQ19}          & (0.3,30,20) & (0.35,10,10) & (0.3,15,20) & (0.4,30,20) & (0.45,20,25)  & (0.45,20,25)   & (0.45,15,10)  & (0.35,10,10) \\
		\hline
		BMF \cite{FleishmanDC03}     & 10 & 35 & 35 & 35 & 35 & 35 & 35 & 35\\
		BNF (local) \cite{Zheng11}   & (20,0.55,10) & (20,0.5,10) & (30,0.45,10) & (30,0.45,20) & (30,0.45,10) & (20,0.35,10)  & (20,0.35,10)  & (20,0.35,10)  \\	
		GNF \cite{cgf/ZhangDZBL15}   & (40,0.3,30) & (10,0.45,10) & (20,0.35,10) & (30,0.45,20) & (30,0.4,10) & (20,0.35,10) & (20,0.35,10) & (20,0.35,10)\\	
		GGNF \cite{zhao2019graph}    & $--$ & $--$  & (Default) & $--$ & $--$ & $--$ & $--$ & $--$ \\
		ROFI \cite{YadavRP19}        & (0.65, 0.25, 100) & (0.7, 0.1, 30) & (0.65, 0.3, 100) & (0.65, 0.35, 30) & (0.6, 0.3, 100) & (0.53, 0.42, 10) & (0.64, 0.35, 20) & (0., 0.42, 10)\\	
		Lu et al. \cite{lu2015robust}     & $--$ & $--$ & $--$ & $--$ & $--$ & $--$ & $--$ & $--$ \\
		SDF \cite{ZhangDHPQL19}      & (0.5,1.2,1.5,0.325) & (1.0,1.5,1.5,0.75) & (1.0,1.2,1.5,0.3) & (0.25,1.5,1.5,0.3) & (1.0,1.5,1.5,0.3)  & (0.25,1.5,1.5,0.3)  & (0.25,1.5,1.5,0.3)  & (0.25,1.5,1.5,0.3) \\ 
		\hline
		CNR \cite{WangLT16}          & (Default) & (Default) & (Default) & (Default) & (Default) & (Default) & (Default) & (Default) \\
		DGRMD \cite{WangHWWXQ19}     & (Default) & (Default) & (Default) & (Default) & (Default) & (Default) & (Default) & (Default)\\
		NormalNet \cite{zhao2019normalnet} & $--$ & $--$ & $--$ & $--$ & $--$ & $--$ & $--$ & $--$\\
		NormalF-Net \cite{li2020normalf}   & (Default) & (Default) & (Default) & (Default) & (Default) & (Default) & (Default) & (Default) \\
		\hline
	\end{tabular}
\end{table*}

\begin{table*}[]
	\centering
	\caption{An overview of the methods for mesh denoising.}
	\label{tab:comp:description}
	\begin{tabular}{|c|c|c|c|c|c|c|}
		\hline
		\textbf{Category} & \textbf{Methods} & \textbf{A1}  & \textbf{A2} & \textbf{A3} & \textbf{A4} & \textbf{Detailed performance}\\ \hline
		\multirow{8}*{I} 
		& He et al. \cite{he2013mesh}          & \xmark  & \checkmark  & \xmark & \xmark & Sharp feature preservation, but time-consuming \\
		& Cheng et al. \cite{cheng2014feature} & \xmark  & \checkmark  & \xmark & \xmark & Better sharp feature preservation than \cite{he2013mesh} \\
		& Zhao et al. \cite{zhao2018robust}    & \xmark  & \checkmark  & \xmark & \xmark & Measuring the sparsity by vertex and normals \\
		& Liu et al. \cite{LiuLZW19}           & \checkmark  & \checkmark & \xmark & \xmark & Handling both sharp features and smoothly curved regions \\
		& Pan et al. \cite{pan2020hlo}         & \checkmark & \xmark & \xmark & \xmark & Handling meshes with high noise \\
		& Li et al. \cite{LiZFH18}             & \checkmark  & \xmark & \checkmark & \xmark & Fine detail preservation \\
		& Wei et al.  \cite{WeiHXLWQ19}   & \checkmark & \checkmark & \checkmark  & \xmark & Preserving both sharp and fine details \\
		& Chen et al. \cite{ChenHRXQGWW19}    & \xmark  & \checkmark & \xmark  & \checkmark & Smoothing geometric features \\
		\hline
		\multirow{5}*{II} 
		& Fleishman et al. \cite{FleishmanDC03}     & \xmark  & \checkmark & \xmark & \xmark & Fast and easy to implement \\
		& Jones et al. \cite{Jones03}               & \checkmark   & \checkmark & \xmark & \xmark & Non-iterative \\
		& Taubin et al. \cite{taubin1995signal}     & \xmark  & \checkmark & \xmark & \xmark & Isotropic and suitable for any topological surface\\
		& Desbrun et al. \cite{desbrun1999implicit} & \xmark  & \xmark & \checkmark  & \xmark & Isotropic \\	
		& Yagou et al. \cite{YagouOB03}             & \xmark  & \checkmark & \xmark & \xmark & Avoiding over-smoothing \\	
		\hline
		\multirow{8}*{III} 
		& Yagou et al. \cite{yagou2002mesh}     & \checkmark  & \xmark & \xmark & \xmark & May cause over-smoothing \\
		& Shen et al. \cite{shen2004fuzzy}      & \checkmark  & \xmark  & \xmark & \xmark & Easy to implement \\
		& Sun et al. \cite{sun2007fast}         & \checkmark  & \xmark & \xmark & \xmark & Efficient and effective vertex updating \\
		& Lee et al. \cite{LeeW05}              & \checkmark  & \xmark & \xmark  & \xmark & Avoiding over-smoothing \\	
		& Zheng et al. \cite{Zheng11}           & \checkmark  & \checkmark & \checkmark   & \xmark & Providing local and global bilateral normal filters \\	
		& Zhang et al. \cite{cgf/ZhangDZBL15}   & \checkmark  & \checkmark & \checkmark & \xmark & Introducing reliable guidance to preserve sharp features \\	
		& Yadav et al. \cite{YadavRP19}         & \checkmark  & \checkmark & \checkmark & \checkmark & High fidelity of the smoothed results \\	
		& Bian et al. \cite{BianT11}           & \checkmark   & \checkmark & \checkmark & \xmark & Robust to irregular sampling \\	
		\hline
		\multirow{8}*{IV} 
		& Wang et al. \cite{wang2012cascaded}  & \checkmark  & \checkmark  & \xmark & \xmark & Well preserving fine details  \\
		& Wei et al. \cite{wei2014bi}          & \checkmark  & \checkmark  & \xmark & \checkmark & Preventing vertex drifts \\
		& Lu et al. \cite{lu2015robust}        & \checkmark  & \checkmark  & \xmark & \xmark & Well preserving fine details \\
		& Yadav et al. \cite{YadavRP18}        & \xmark  & \checkmark & \xmark & \xmark & Speeding up convergence with quadratic optimization \\
		& Fan et al. \cite{fan2009robust}      & \checkmark & \checkmark & \checkmark  & \checkmark & Filtering according to vertex type\\	
		& Zhu et al. \cite{zhu2013coarse}      & \checkmark & \checkmark & \checkmark & \xmark & Providing accurate normals by isotropic neighborhood \\	
		& Wang et al. \cite{WangFLTLG15}       & \xmark  & \checkmark & \checkmark  & \checkmark & Avoiding mesh facet flipping \\
		& Zhang et al. \cite{ZhangDHPQL19}     & \xmark & \checkmark & \checkmark   & \checkmark & Fast to remove geometric details \\												
		\hline
		\multirow{4}*{V}  
		& Wang et al. \cite{WangLT16}           & \checkmark  & \checkmark & \checkmark  & \xmark & The first learning-based mesh smoothing method \\
		& Wang et al. \cite{WangHWWXQ19}        & \checkmark  & \checkmark & \checkmark  & \checkmark & Can preserve different scales of features \\
		& Wei et al.  \cite{WeiGHXZKWQ19}       & \checkmark  & \checkmark & \checkmark  & \xmark & Can recover lost geometry during mesh
		denoising\\
		& Zhao et al. \cite{zhao2019normalnet}  & \checkmark  & \checkmark & \checkmark  & \checkmark & Precise guidance, better effect than traditional methods \\
		& Li et al. \cite{li2020normalf}        & \checkmark & \checkmark & \checkmark  & \checkmark & Better effect than traditional methods, but time-consuming \\
		\hline
	\end{tabular}
\end{table*}

\begin{table*}[]
	\centering
	\setlength{\tabcolsep}{1.2mm}
	\caption{Quantitative comparisons of the representative mesh denoising methods.}
	\label{tab:comp:error}
	\begin{tabular}{|c|cc|cc|cc|cc|cc|cc|cc|cc|}
		\hline
		\multirow{2}{*}{\textbf{Methods}} & \multicolumn{2}{c|}{Block} & \multicolumn{2}{c|}{Child} & \multicolumn{2}{c|}{Fandisk} & \multicolumn{2}{c|}{Boy} & \multicolumn{2}{c|}{Cone}  & \multicolumn{2}{c|}{Scene-1} & \multicolumn{2}{c|}{Scene-2} & \multicolumn{2}{c|}{Scene-3}\\ \cline{2-17}
		&$\theta$ & $d$ & $\theta$ & $d$ & $\theta$ & $d$ & $\theta$ & $d$ & $\theta$ & $d$ & $\theta$ & $d$ & $\theta$ & $d$& $\theta$ & $d$\\ \hline
		$L_0$ \cite{he2013mesh}          & 5.4 & 0.121 & 11.12 & 0.0018 & 11.37 & 0.037 & 9.24 & 1.191 & 9.36 & 0.854 & 3.57 & 0.338 & 5.02 & 0.154 & 7.72 & 0.215\\
		HOF \cite{LiuLZW19}   & 5.36 & 0.099 & 7.97 & 0.0011 & 3.11 & 0.022 & 9.10 & 1.242 & 9.02 & 0.997 & $--$ & $--$ & $--$ & $--$ & $--$ & $--$\\
		HLO \cite{pan2020hlo}  & 13.79 & 0.197 & 8.25 & 0.0011 & 11.66 & 0.023 & 9.27 & 1.197 & 9.29 & 1.417 & 4.56 & 0.333 & 7.04 & 0.219 & 10.23& 0.298\\
		NLLR \cite{LiZFH18}    & 11.31 & 0.109 & 6.56 & 0.0006 & 8.59 & 0.013 & 10.4 & 1.196 & 11.51 & 1.126  & 3.19 & 0.335 & $--$ & $--$ & $--$ & $--$\\
		PCF  \cite{WeiHXLWQ19} & 4.56 &  \textbf{0.061} & \textbf{5.08} & 0.0007 & 3.00 & 0.009  & \textbf{8.92} & \textbf{1.173} & \textbf{8.19} & 0.792 & 3.12 & 0.338 & 4.85 & 0.133 & 8.54 & 0.234\\
		\hline
		BMF \cite{FleishmanDC03}     & 15.03 & 0.165  & 9.92 & 0.0010 & 13.12 & 0.038 & 12.13 & 1.215 & 12.50 & 0.961 & 4.15 & 0.345 & 8.43 & 0.192 & 10.85 & 0.266 \\
		BNF (local) \cite{Zheng11}   & 11.73 & 0.123 & 8.19 & 0.0008 & 7.08 & 0.014 & 10.52 & 1.193 & 14.43 & 1.037 & 3.42 & 0.332 & 5.18 & 0.140 & 8.46 & 0.196\\	
		GNF \cite{cgf/ZhangDZBL15}   & 4.29 & 0.067 & 8.56 & 0.0008 & 2.93 & 0.007 & 9.38 & 1.192 & 11.40 & 0.934 & 3.13 & 0.332 & 4.47 & 0.132 & 7.29 & 0.195\\	
		GGNF \cite{zhao2019graph}    & $--$ & $--$ & $--$ & $--$ & 2.3 & 0.006 & 9.11 & 1.187 & 9.02 & 0.827 &  $--$ &  $--$ & $--$ & $--$ & $--$ & $--$\\
		ROFI \cite{YadavRP19}        & 5.18 & 0.090 & 7.26 & 0.0007 & 4.35 & 0.013 & 10.14 & 1.181 & 10.97 & 0.894 & 3.48 & 0.032 & 4.32 & 0.138 & 7.30 & \textbf{0.180}\\	
		Lu et al. \cite{lu2015robust}     & $--$ & $--$ & $--$ & $--$ & 3.28 & 0.014 & $--$ & $--$ & $--$ & $--$ & $--$ & $--$ & $--$ & $--$ & $--$ & $--$ \\
		SDF \cite{ZhangDHPQL19}      & 10.71 & 0.273 & 15.34 & 0.0027 & 4.79 & 0.028 & 9.08 & 1.198 & 10.19 & 1.193 & 2.93 & 0.336 & 7.28 & 0.196 & 7.91 & 0.251 \\										
		\hline
		CNR \cite{WangLT16}          & 12.72 & 0.147 & 7.93 & 0.0007 & 2.7 & 0.007 & 9.45 & 1.189 & 8.50 & \textbf{0.791} & \textbf{2.63} & 0.334 & \textbf{3.77} & \textbf{0.122} & \textbf{6.50} & 0.184 \\
		DGRMD \cite{WangHWWXQ19}     & \textbf{3.75} & 0.088 & \textbf{5.08} & \textbf{0.0004} & \textbf{1.87} & \textbf{0.005} & 9.28 & 1.187 & 8.85 & 0.794  & 3.62 & 0.333& 4.32 & 0.128 & 6.78 & 0.190\\
		NormalNet \cite{zhao2019normalnet} & $--$ & $--$ & 9.33 & 0.0008 & 2.13 & 0.007 & 8.94 & 1.182 & 8.32 & 0.814 & $--$ & $--$ & $--$ & $--$ & $--$ & $--$ \\
		NormalF-Net \cite{li2020normalf} & 11.99 & 0.113 & 7.17 & 0.0007 & 3.00 & 0.013 & 9.92 & 1.199 & 9.98 & 0.896 & 3.40 & 0.332 & 5.44 & 0.148 & 7.54 & 0.206\\
		\hline
	\end{tabular}
\end{table*}

\begin{figure*}
	\centering
	\subfigure[Noisy input (4$\%$ Gaussian noise)]{
		\includegraphics[width=1.3in]{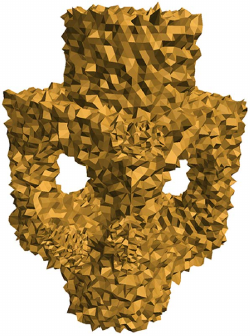}}
	\subfigure[Ground truth]{
		\includegraphics[width=1.3in]{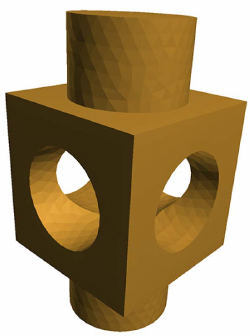}}
	\subfigure[BMF \cite{FleishmanDC03}]{
		\includegraphics[width=1.3in]{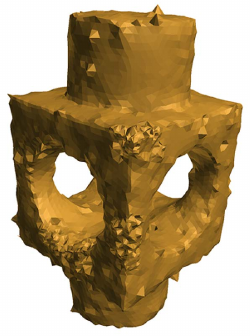}}
	\subfigure[BNF (local) \cite{Zheng11}]{
		\includegraphics[width=1.3in]{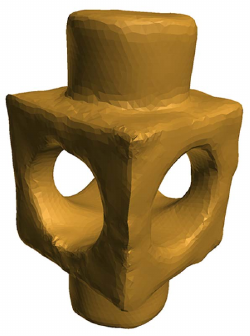}}
	\subfigure[GNF \cite{cgf/ZhangDZBL15}]{
		\includegraphics[width=1.3in]{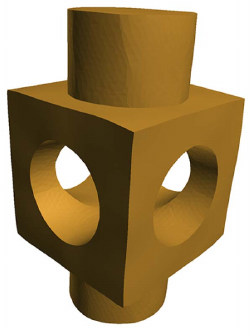}}
	\subfigure[$L_0$ \cite{he2013mesh} ]{
		\includegraphics[width=1.3in]{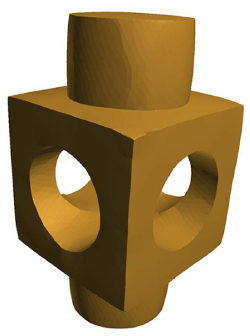}}
	\subfigure[NLLR \cite{LiZFH18}]{
		\includegraphics[width=1.3in]{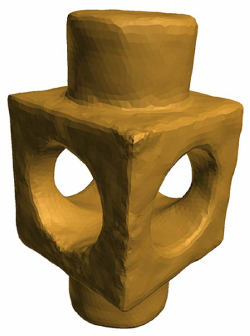}}
	\subfigure[HOF \cite{LiuLZW19}]{
		\includegraphics[width=1.3in]{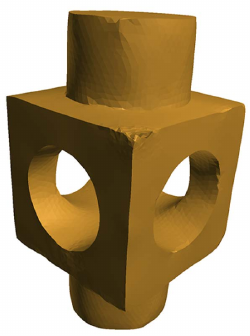}}
	\subfigure[ROFI \cite{YadavRP19}]{
		\includegraphics[width=1.3in]{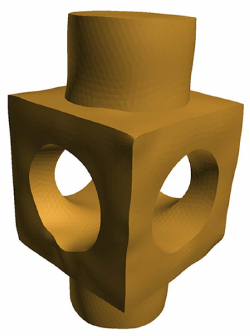}}
	\subfigure[HLO \cite{pan2020hlo}]{
		\includegraphics[width=1.3in]{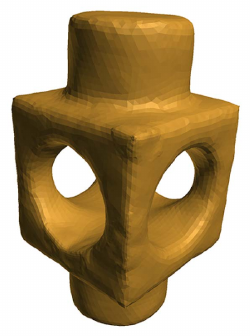}}
	\subfigure[SDF \cite{ZhangDHPQL19}]{
		\includegraphics[width=1.3in]{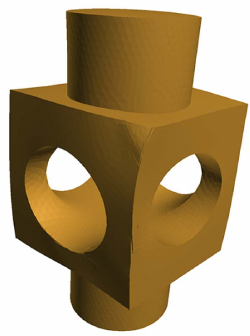}}
	\subfigure[PCF \cite{WeiHXLWQ19}]{
		\includegraphics[width=1.3in]{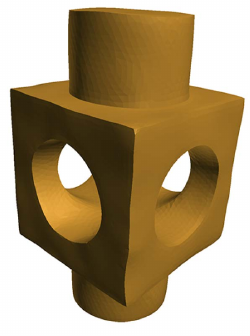}}
	\subfigure[CNR \cite{WangLT16}]{
		\includegraphics[width=1.3in]{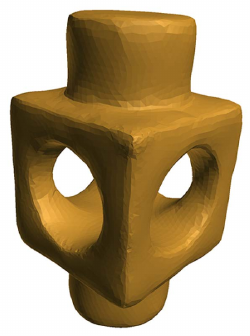}}
	\subfigure[DGRMD \cite{WangHWWXQ19}]{
		\includegraphics[width=1.3in]{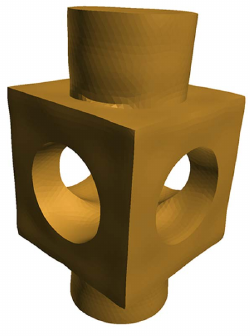}}
	\subfigure[NormalF-Net \cite{li2020normalf}]{
		\includegraphics[width=1.3in]{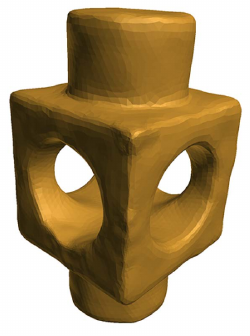}}
	\caption{Comparison of the state-of-the-art mesh denoising algorithms. Visually speaking, GNF \cite{cgf/ZhangDZBL15} produces the most pleasing result.}
	\label{block}
\end{figure*}

\begin{figure*}
	\centering
	\subfigure[Noisy input (3$\%$ Gaussian noise)]{
		\includegraphics[width=1.3in]{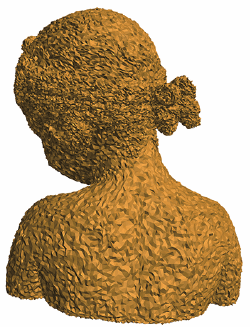}}
	\subfigure[Ground truth]{
		\includegraphics[width=1.3in]{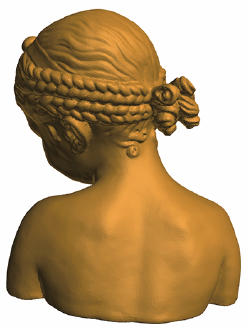}}
	\subfigure[BMF \cite{FleishmanDC03}]{
		\includegraphics[width=1.3in]{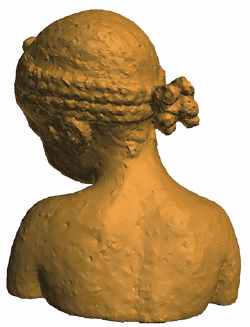}}
	\subfigure[BNF (local) \cite{Zheng11}]{
		\includegraphics[width=1.3in]{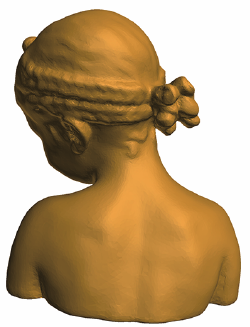}}
	\subfigure[GNF \cite{cgf/ZhangDZBL15}]{
		\includegraphics[width=1.3in]{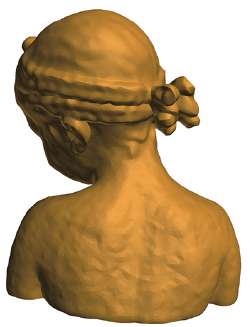}}
	\subfigure[$L_0$ \cite{he2013mesh} ]{
		\includegraphics[width=1.3in]{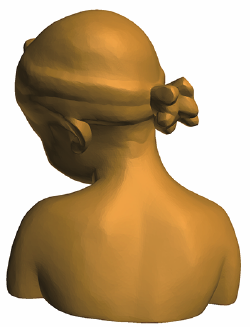}}
	\subfigure[NLLR \cite{LiZFH18}]{
		\includegraphics[width=1.3in]{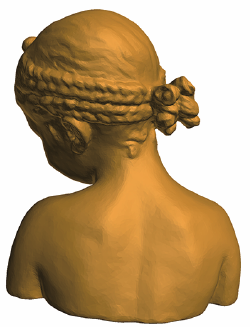}}
	\subfigure[HOF \cite{LiuLZW19}]{
		\includegraphics[width=1.3in]{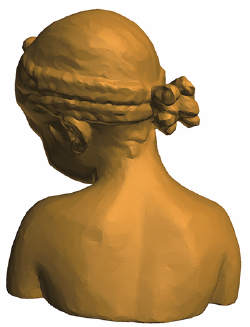}}
	\subfigure[ROFI \cite{YadavRP19}]{
		\includegraphics[width=1.3in]{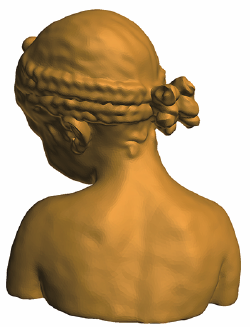}}
	\subfigure[HLO \cite{pan2020hlo}]{
		\includegraphics[width=1.3in]{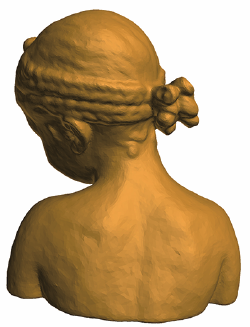}}
	\subfigure[SDF \cite{ZhangDHPQL19}]{
		\includegraphics[width=1.3in]{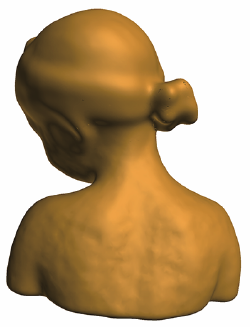}}
	\subfigure[PCF \cite{WeiHXLWQ19}]{
		\includegraphics[width=1.3in]{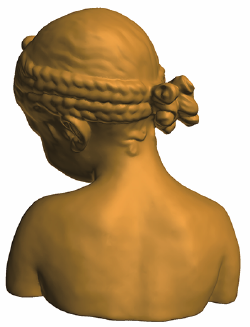}}
	\subfigure[CNR \cite{WangLT16}]{
		\includegraphics[width=1.3in]{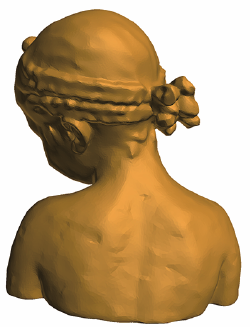}}
	\subfigure[DGRMD \cite{WangHWWXQ19}]{
		\includegraphics[width=1.3in]{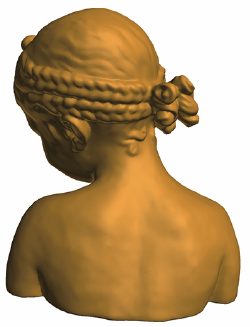}}
	\subfigure[NormalF-Net \cite{li2020normalf}]{
		\includegraphics[width=1.3in]{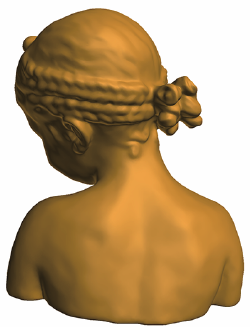}}
	\caption{Comparison of the state-of-the-art mesh denoising algorithms. We find that DGRMD \cite{WangHWWXQ19} better preserves the geometric features. }
	\label{child}
\end{figure*}

\begin{figure*}
	\centering
	\subfigure[Noisy input (3$\%$ Gaussian noise)]{
		\includegraphics[width=1.1in]{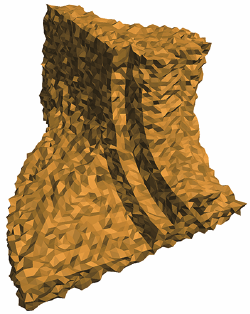}}
	\subfigure[Ground truth]{
		\includegraphics[width=1.1in]{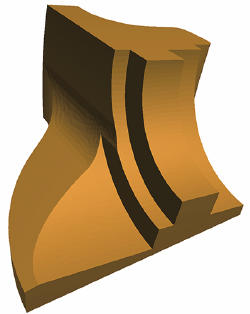}}
	\subfigure[BMF \cite{FleishmanDC03}]{
		\includegraphics[width=1.1in]{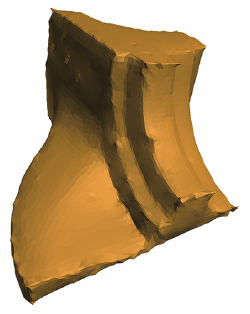}}
	\subfigure[BNF \cite{Zheng11}]{
		\includegraphics[width=1.1in]{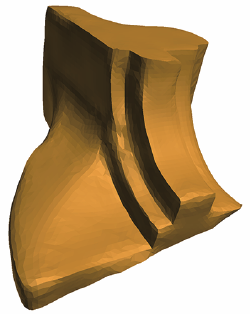}}
	\subfigure[GNF \cite{cgf/ZhangDZBL15}]{
		\includegraphics[width=1.1in]{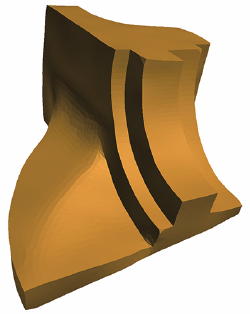}}
	\subfigure[GGNF \cite{zhao2019graph}]{
		\includegraphics[width=1.1in]{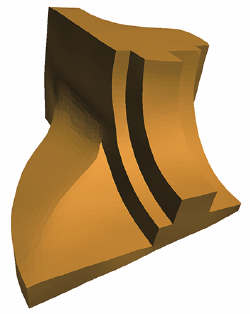}}
	\subfigure[Lu et al. \cite{lu2015robust}]{
		\includegraphics[width=1.1in]{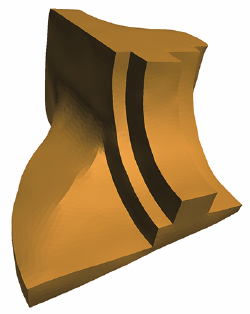}}
	\subfigure[$L_0$ \cite{he2013mesh} ]{
		\includegraphics[width=1.1in]{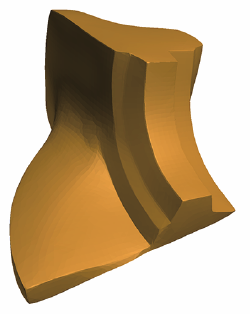}}
	\subfigure[NLLR \cite{LiZFH18}]{
		\includegraphics[width=1.1in]{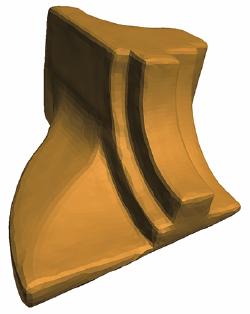}}
	\subfigure[HOF \cite{LiuLZW19}]{
		\includegraphics[width=1.1in]{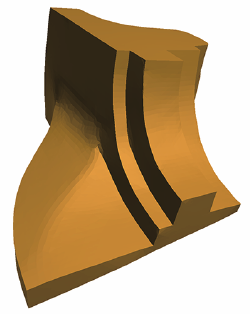}}
	\subfigure[ROFI \cite{YadavRP19}]{
		\includegraphics[width=1.1in]{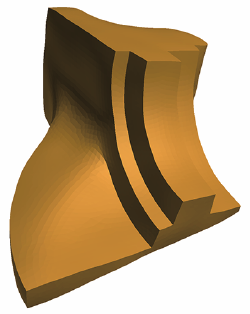}}
	\subfigure[HLO \cite{pan2020hlo}]{
		\includegraphics[width=1.1in]{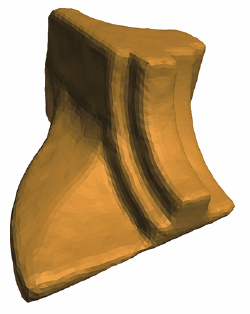}}
	\subfigure[SDF \cite{ZhangDHPQL19}]{
		\includegraphics[width=1.1in]{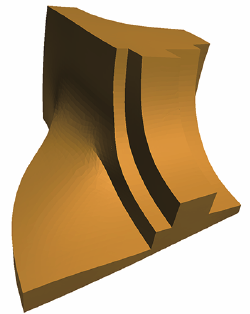}}
	\subfigure[PCF \cite{WeiHXLWQ19}]{
		\includegraphics[width=1.1in]{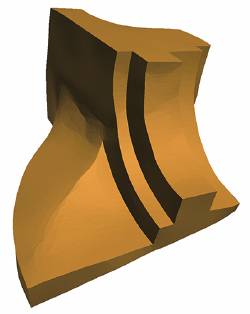}}
	\subfigure[CNR \cite{WangLT16}]{
		\includegraphics[width=1.1in]{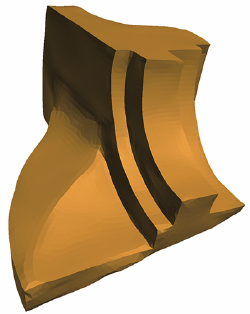}}
	\subfigure[NormalNet \cite{zhao2019normalnet}]{
		\includegraphics[width=1.1in]{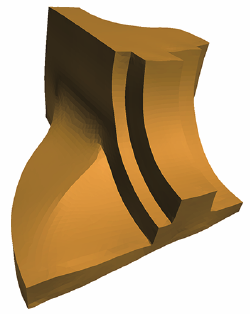}}
	\subfigure[DGRMD \cite{WangHWWXQ19}]{
		\includegraphics[width=1.1in]{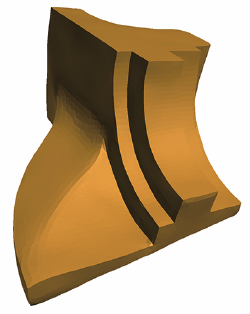}}
	\subfigure[NormalF-Net \cite{li2020normalf}]{
		\includegraphics[width=1.1in]{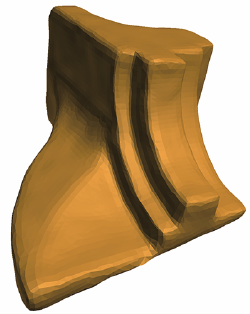}}
	\caption{Comparison of the state-of-the-art mesh denoising algorithms. We find global optimization methods \cite{he2013mesh,YadavRP19} cannot deal well with shallow features. }
	\label{fandisk}
\end{figure*}

\begin{figure*}
	\centering
	\subfigure[Noisy input (real scan from \cite{WangLT16})]{
		\includegraphics[width=1.1in]{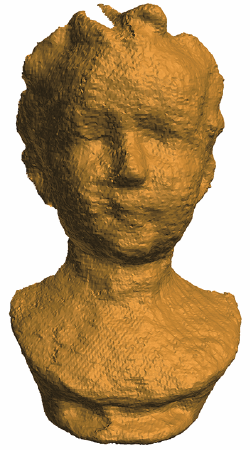}}
	\subfigure[Ground truth]{
		\includegraphics[width=1.1in]{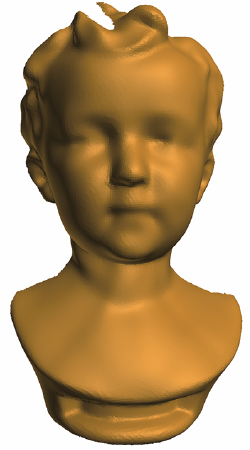}}
	\subfigure[BF \cite{FleishmanDC03}]{
		\includegraphics[width=1.1in]{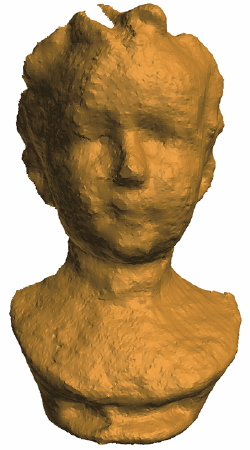}}
	\subfigure[BNF \cite{Zheng11}]{
		\includegraphics[width=1.1in]{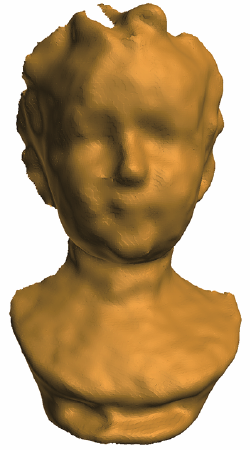}}
	\subfigure[GNF \cite{cgf/ZhangDZBL15}]{
		\includegraphics[width=1.1in]{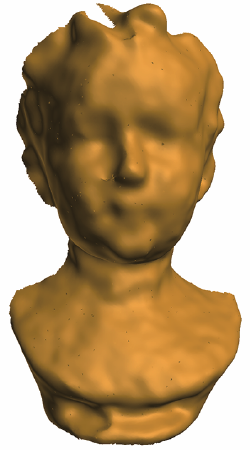}}
	\subfigure[GGNF \cite{zhao2019graph}]{
		\includegraphics[width=1.1in]{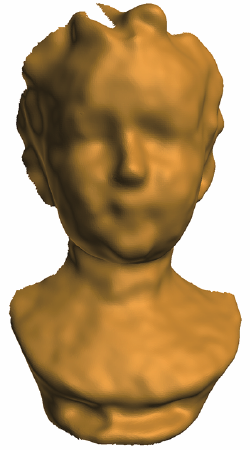}}
	\subfigure[$L_0$ \cite{he2013mesh} ]{
		\includegraphics[width=1.1in]{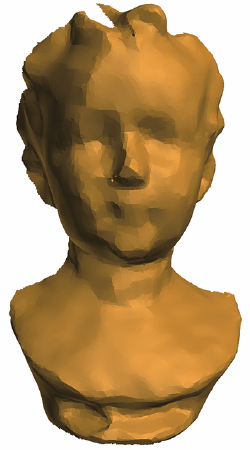}}
	\subfigure[NLLR \cite{LiZFH18}]{
		\includegraphics[width=1.1in]{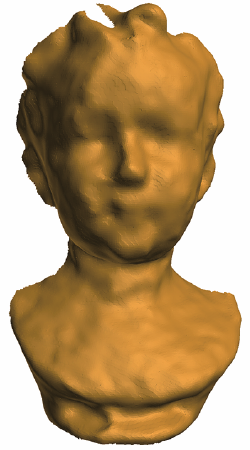}}
	\subfigure[HOF \cite{LiuLZW19}]{
		\includegraphics[width=1.1in]{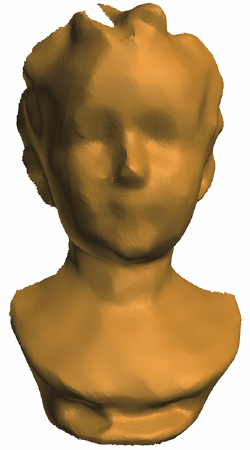}}
	\subfigure[ROFI \cite{YadavRP19}]{
		\includegraphics[width=1.1in]{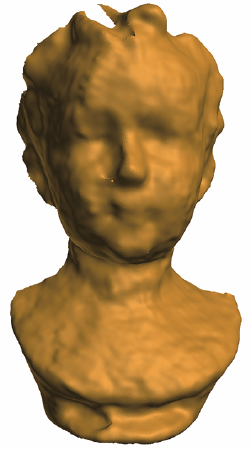}}
	\subfigure[HLO \cite{pan2020hlo}]{
		\includegraphics[width=1.1in]{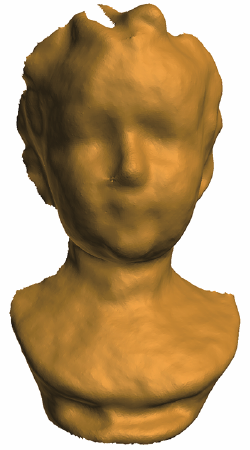}}
	\subfigure[SDF \cite{ZhangDHPQL19}]{
		\includegraphics[width=1.1in]{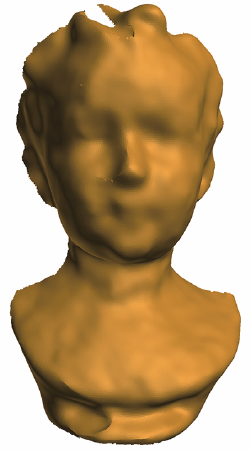}}
	\subfigure[PCF \cite{WeiHXLWQ19}]{
		\includegraphics[width=1.1in]{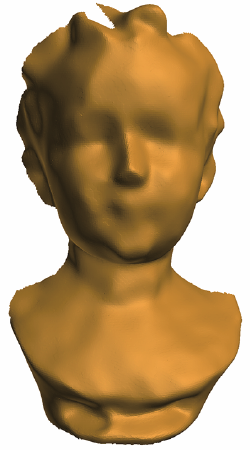}}
	\subfigure[CNR \cite{WangLT16}]{
		\includegraphics[width=1.1in]{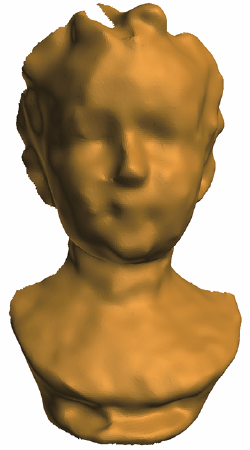}}
	\subfigure[NormalNet \cite{zhao2019normalnet}]{
		\includegraphics[width=1.1in]{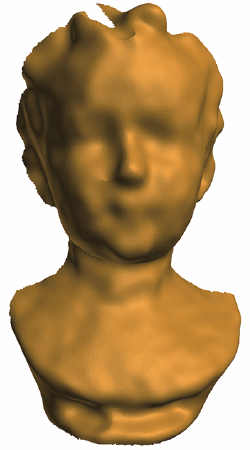}}
	\subfigure[DGRMD \cite{WangHWWXQ19}]{
		\includegraphics[width=1.1in]{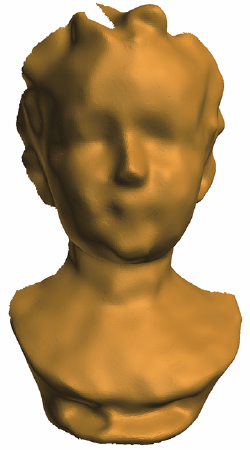}}
	\subfigure[NormalF-Net \cite{li2020normalf}]{
		\includegraphics[width=1.1in]{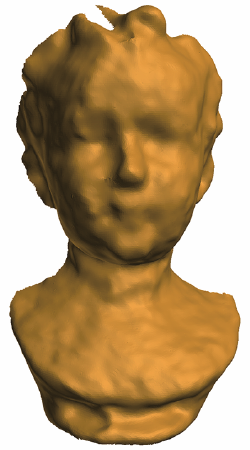}}
	\caption{Comparison of the state-of-the-art mesh denoising algorithms. We find that DGRMD \cite{WangHWWXQ19} better preserves geometric features. }
	\label{boy01}
\end{figure*}

\begin{figure*}
	\centering
	\subfigure[Noisy input (real scan from \cite{WangLT16})]{
		\includegraphics[width=1.1in]{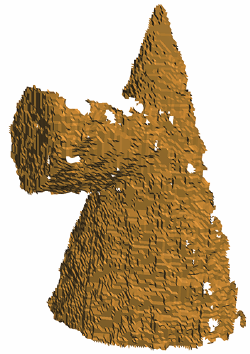}}
	\subfigure[Ground truth]{
		\includegraphics[width=1.1in]{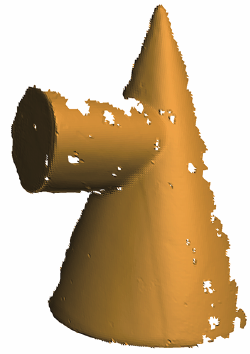}}
	\subfigure[BMF \cite{FleishmanDC03}]{
		\includegraphics[width=1.1in]{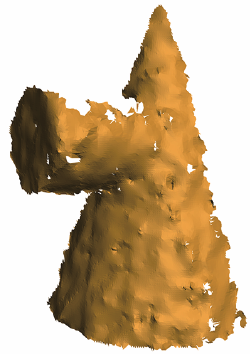}}
	\subfigure[BNF \cite{Zheng11}]{
		\includegraphics[width=1.1in]{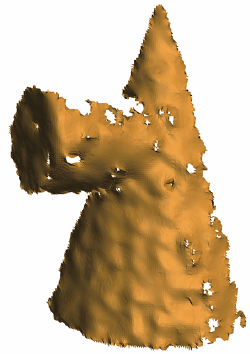}}
	\subfigure[GNF \cite{cgf/ZhangDZBL15}]{
		\includegraphics[width=1.1in]{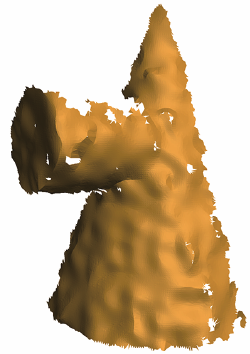}}
	\subfigure[GGNF \cite{zhao2019graph}]{
		\includegraphics[width=1.1in]{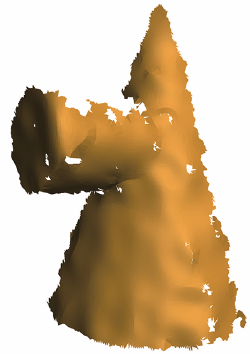}}
	\subfigure[$L_0$ \cite{he2013mesh} ]{
		\includegraphics[width=1.1in]{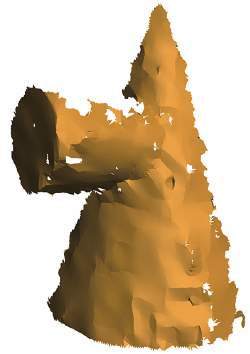}}
	\subfigure[NLLR \cite{LiZFH18}]{
		\includegraphics[width=1.1in]{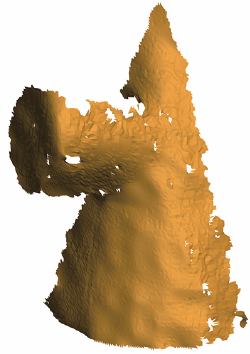}}
	\subfigure[HOF \cite{LiuLZW19}]{
		\includegraphics[width=1.1in]{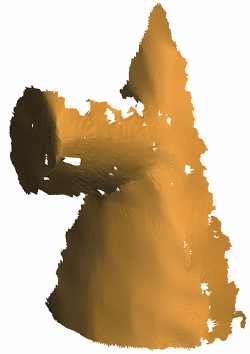}}
	\subfigure[ROFI \cite{YadavRP19}]{
		\includegraphics[width=1.1in]{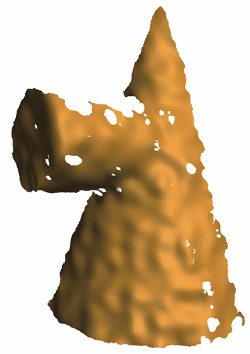}}
	\subfigure[HLO \cite{pan2020hlo}]{
		\includegraphics[width=1.1in]{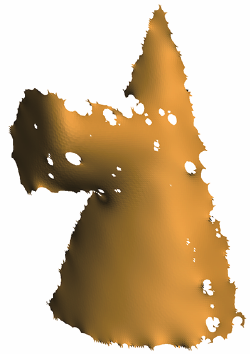}}
	\subfigure[SDF \cite{ZhangDHPQL19}]{
		\includegraphics[width=1.1in]{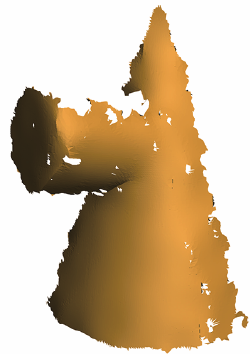}}
	\subfigure[PCF \cite{WeiHXLWQ19}]{
		\includegraphics[width=1.1in]{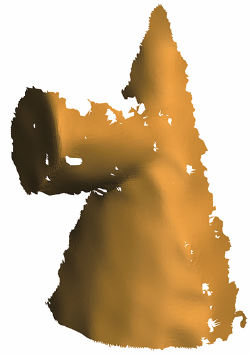}}
	\subfigure[CNR \cite{WangLT16}]{
		\includegraphics[width=1.1in]{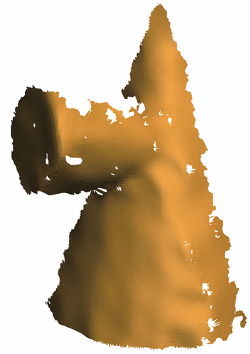}}
	\subfigure[NormalNet \cite{zhao2019normalnet}]{
		\includegraphics[width=1.1in]{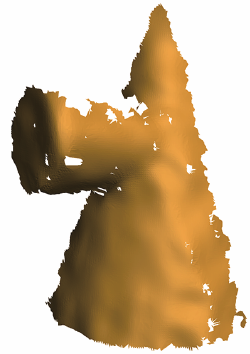}}
	\subfigure[DGRMD \cite{WangHWWXQ19}]{
		\includegraphics[width=1.1in]{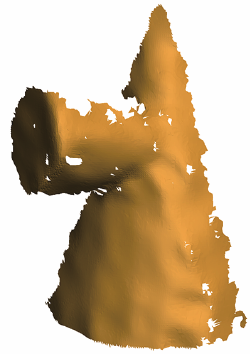}}
	\subfigure[NormalF-Net \cite{li2020normalf}]{
		\includegraphics[width=1.1in]{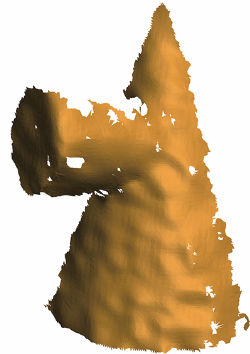}}
	\caption{Comparison of the state-of-the-art mesh denoising algorithms. We find that learning-based methods can not completely smooth out stepping noise.}
	\label{cone04}
\end{figure*}

\vfill


\vfill

\end{document}